\documentclass[times,final]{elsarticle}

\usepackage[english]{babel}

\usepackage[letterpaper,top=2cm,bottom=2cm,left=3cm,right=3cm,marginparwidth=1.75cm]{geometry}

\usepackage{amsmath}
\usepackage{graphicx}
\usepackage[colorlinks=true, allcolors=blue]{hyperref}
\usepackage{subcaption}

\usepackage{array}
\newcolumntype{C}[1]{>{\centering\arraybackslash}p{#1}}

\usepackage[table]{xcolor}
\usepackage{lineno}
\usepackage{setspace}

\begin{document}

\begin{frontmatter}
\title{A reinforcement learning strategy for p-adaptation in high order solvers}
\author[1]{David Huergo}
\author[1,2]{Gonzalo Rubio}
\author[1,2]{Esteban Ferrer}
\cortext[cor1]{Corresponding author}
\ead{esteban.ferrer@upm.es}

\address[1]{ETSIAE-UPM-School of Aeronautics, Universidad Politécnica de Madrid, Plaza Cardenal Cisneros 3, E-28040 Madrid, Spain}
\address[2]{Center for Computational Simulation, Universidad Politécnica de Madrid, Campus de Montegancedo, Boadilla del Monte, 28660 Madrid, Spain}


\begin{abstract}
Reinforcement learning (RL) has emerged as a promising approach to automating decision processes. This paper explores the application of RL techniques to optimise the polynomial order in the computational mesh when using high-order solvers. Mesh adaptation plays a crucial role in improving the efficiency of numerical simulations by improving accuracy while reducing the cost. Here, actor-critic RL models based on  Proximal Policy Optimization offer a data-driven approach for agents to learn optimal mesh modifications based on evolving conditions.
%

The paper provides a strategy for p-adaptation in high-order solvers and includes insights into the main aspects of RL-based mesh adaptation, including the formulation of appropriate reward structures and the interaction between the RL agent and the simulation environment. We discuss the impact of RL-based mesh p-adaptation on computational efficiency and accuracy. We test the RL p-adaptation strategy on a 1D inviscid Burgers' equation to demonstrate the effectiveness of the strategy. The RL strategy reduces the computational cost and improves accuracy over uniform adaptation, while minimising human intervention.  

\end{abstract}

\end{frontmatter}


\onehalfspacing

\section{Introduction}

In recent years, the field of machine learning has witnessed remarkable advancements, enabling machines to learn and adapt to complex tasks with increasing autonomy. Of particular interest are advances related to fluid mechanics and the related solution of partial differential equations \cite{brunton_kutz_2019,GARNIER2021104973,vinuesa2022enhancing,Vinuesa_Brunton,brenner2019perspective,sole_review}. 
Reinforcement learning (RL), a prominent subfield of machine learning, has emerged as a powerful framework to enable intelligent decision making in dynamic and uncertain environments. RL encompasses algorithms and methodologies that enable agents to learn optimal strategies by interacting with an environment and receiving feedback in the form of rewards or penalties.

Reinforcement learning has garnered significant attention across various domains, ranging from robotics and game-playing to finance and healthcare. Its success lies in its ability to handle sequential decision-making problems, where an agent learns to make optimal decisions by taking into account the current state of the environment, selecting appropriate actions, and observing the resulting rewards or consequences. This iterative process of learning from interactions allows RL agents to navigate through complex environments, optimise their behaviours, and achieve predefined objectives.

 In the field of computational fluid dynamics (CFD), which deals with the numerical simulation of fluid flow, RL techniques have shown promise for flow control (e.g., drag reduction), shape optimisation, and turbulence modelling \cite{GARNIER2021104973,10.1063/5.0128446}. 
RL is still in its infancy when combined with numerical techniques. In particular, only few publications have studied RL to control mesh adaptation in CFD \cite{yang2023reinforcement,wu2023learning,10.5555/3545946.3598614,10.1063/5.0138039,freymuth2023swarm} and concluded that RL policies have the potential to outperform adaptation based on error estimators. Let us note that all the previous works have considered low-order numerical methods, and to the authors' knowledge there is no RL strategy for high-order p-adaptation, which is the topic of this work.
Mesh adaptation plays a crucial role in numerical methods, as it allows for the refinement or coarsening of the computational mesh based on the solution smoothness and the computational cost. Traditionally, mesh adaptation has relied on heuristics or manual intervention, which can be time consuming and limited in their ability to capture complex flow phenomena. Automatic mesh adaptation algorithms require a certain criteria to determine the regions that require adjustment. These criteria can be classified into three distinct categories, as discussed in \cite{naddeiComparisonRefinementIndicators2019}: feature-based indicators \cite{ntoukas2021free, ntoukas2022entropy,li2020p,kamkar2011feature}, local error-based indicators \cite{rueda2019p,de2018flow,basile2022unstructured}, and goal-oriented indicators \cite{laskowski2022functional,dwight2008goal,burgess2011hp}.
Feature-based indicators are derived from the physical properties of the flow (e.g., norm of the gradient), and although these methods are relatively inexpensive and straightforward to implement, they often lack robustness and their effectiveness can vary depending on the specific case.
The second category, local error-based indicators, identifies regions that require refinement based on high values of quantified error. Implementing this approach involves obtaining solutions at different levels of mesh refinement, which can be computationally expensive until an appropriate numerical setup is achieved. Finally goal-oriented indicators evaluate the error contribution associated with a specific target functional, often relying on adjoint-based techniques. However, this approach is costly as it necessitates solving a new system of equations.

RL offers a novel approach to automate and optimise the mesh adaptation process by enabling the solver to learn the most suitable mesh modifications based on the time-evolving conditions.
By integrating reinforcement learning algorithms into simulations, it is possible to train agents to adapt the computational mesh (here increasing/decreasing the polynomial order in a discontinuous Galerkin solver) in response to changes in the solution. The RL agent interacts with the numerical solver, observing the state of the flow, making decisions on mesh modifications (e.g., increasing the polynomial order in certain mesh elements), and receiving rewards based on the resulting accuracy and efficiency of the simulation. In fact, it is possible to decouple the training so that a more efficient decision process is incorporated into the numerical solver. This training considers high-order polynomials and is an essential part of the proposed RL strategy studied in this work. 
Through iterative learning, the RL agent improves its mesh adaptation strategies, ultimately leading to more accurate and efficient simulations.

The application of reinforcement learning in mesh adaptation brings several advantages. First, it reduces the reliance on manual intervention, allowing for automated and adaptive mesh refinement. Secondly, RL-based mesh adaptation can optimise the computational resources by dynamically allocating degrees of freedom (here high order polynomials) in regions of interest and reducing unnecessary refinement in areas with low flow variations. This resource optimisation leads to computational cost savings without compromising the accuracy of the solution.
Moreover, reinforcement learning enables the exploration of unconventional mesh adaptations that might not be intuitively considered by traditional methods. RL agents have the capability to discover non-obvious mesh modifications that enhance the accuracy and robustness of the CFD simulations. 



In this work, we design a strategy for polynomial adaptation in high-order methods using RL. Without loss of generality we restrict the methodology to 1D cases and focus on the design of appropriate rewards and state definitions that lead to efficient RL-adaptation. First, we describe the methodology in section \ref{meth}, including the Proximal Policy Optimisation algorithm, the numerical solution of the PDE and the RL strategy (detailing agent, states, rewards, training and testing). Second, we provide results in section \ref{res} and finalise with conclusions in \ref{conc}.

\section{Methodology}\label{meth}

\subsection{Mathematical model and numerical discretisation}

We use a one-dimensional nonlinear partial differential equation to test our strategy for mesh adaptation. In particular, we use the 1D inviscid Burgers' equation:

\begin{equation}
    \frac{\partial u}{\partial t} + \frac{1}{2}\frac{\partial u^2}{\partial x} = 0.
    \label{eq:Burgers_equation}
\end{equation}
We solve this equation in the domain $x \in [0, 1]$, with periodic boundary conditions and the initial condition:
\begin{equation}
    u(x,0) = 2 + \sin{(2\pi x)}.
    \label{eq:Burgers_initial_condition}
\end{equation}

The analytical solution to the previous equation can be obtained through the method of characteristics \cite{kuo1990semi}, leading to:
\begin{equation}
    u(x,t) = h(x - u t),
    \label{eq:Burgers_characteristics}
\end{equation}
where $h$ is a function that can be computed from the initial condition. In this case, we use the condition defined in Eq. \eqref{eq:Burgers_initial_condition}, the final solution is:
\begin{equation}
    u(x,t) = 2 + \sin{(2\pi (x - u t))}.
    \label{eq:Burgers_analytical_solution}
\end{equation}

Equation \eqref{eq:Burgers_analytical_solution} will be used to measure the performance of our methodology in Section \ref{res}. The solutions of Eq. \eqref{eq:Burgers_analytical_solution} for every pair $(x,t)$ are obtained using a Newton-Raphson method.\\

We discretise the equation using a nodal high-order discontinuous Galerkin spectral element method (DGSEM), see details in \ref{app:dgsem}. 
The method allows for the selection of an arbitrary polynomial order, $p$, in each mesh element, which can be different from the ones selected in the neighbouring elements. This is a key advantage of DG since it combines flexibility and accuracy. The discontinuous nature of the method allows for local refinement and adaptivity, enabling precise resolution of features with high gradients or regions of interest. This adaptivity is particularly beneficial when solving PDEs in complex geometries or when dealing with problems in which the solution undergoes abrupt changes. 


\subsection{Reinforcement learning strategy for p-adaptation}
\begin{figure}[h]
    \centering
    \includegraphics[width=0.7\textwidth]{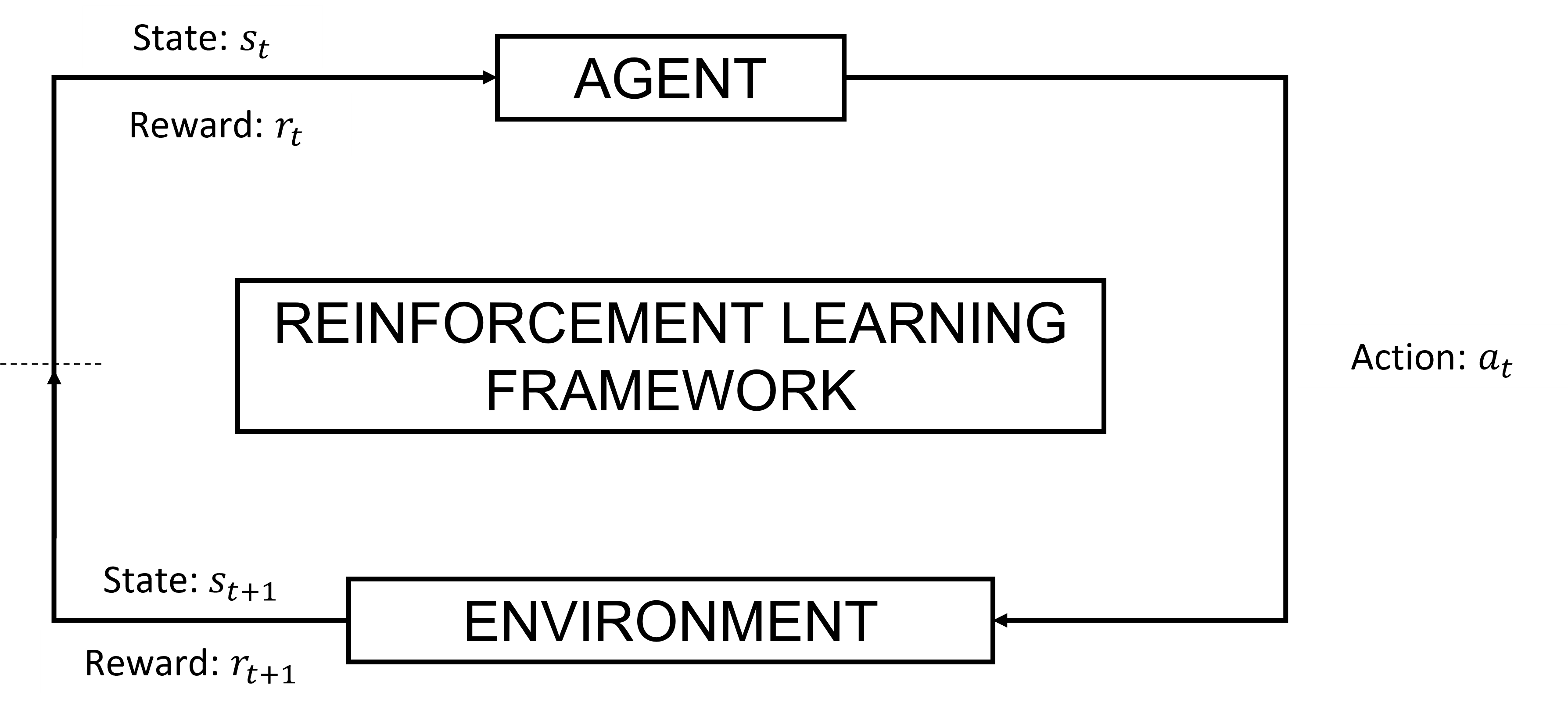}
    \caption{Schematic definition of the RL framework.}
    \label{fig:RL_scheme}
\end{figure}

Reinforcement learning is generally considered a semi-supervised approach, as the agent learns by itself the optimal policy through the interaction with an environment. However, the user must define an objective function to reward the agent when it shows a positive behaviour. Every RL method is based on the scheme shown in Figure~\ref{fig:RL_scheme}.

This framework highlights several points that must be defined for the correct understanding of the proposed methodology:
\begin{enumerate}
    \item \textbf{Agent}: The model which decides the best action to take, in our case the best polynomial order, $p$, in each mesh element. It is made up of two neural networks: the actor and the critic. See more details in section \ref{agent}.
    \item \textbf{Environment}: Each element and polynomial of the computational mesh and the DG scheme. It allows to compute a solution in one element given the polynomial order, $p$.
    \item \textbf{Action}: It is an output for the agent and an input for the environment. Three different actions will be considered: to increase the polynomial order one unit, to decrease the polynomial order one unit, or to keep the polynomial order constant.
    \item \textbf{State}: It is an output for the environment and an input for the agent. It must contain enough information to define the current scenario. See more details in section \ref{state}.
    \item \textbf{Reward}: The user-defined objective function. It should have a higher value when the agent is performing better. See more details in section \ref{reward}.
\end{enumerate}

While the definitions of the environment and the actions are straightforward, the agent, the state, and the reward have to be cautiously designed to obtain good performance.

\subsubsection{Agent definition}\label{agent}
The reinforcement learning model has been trained using the Proximal Policy Optimisation (PPO) algorithm, which allows the agent to learn the optimum policy that maximises the expected cumulative reward for the current state. 
PPO is a reinforcement learning algorithm that has gained significant attention for its effectiveness and stability in training complex decision-making agents \cite{schulman2017proximal}. PPO belongs to the class of policy optimisation algorithms and is designed to strike a balance between exploration and exploitation, enabling agents to learn robust policies.
At its core, PPO aims to maximise the expected cumulative rewards by iteratively updating the policy of an RL agent. 
%
The key idea behind PPO is to define a surrogate objective function that measures the policy improvement. This surrogate objective is optimised through multiple iterations, with each iteration consisting of two steps: sampling trajectories and updating the policy. In the trajectory sampling step, the agent interacts with the environment to collect a batch of experiences. These experiences are then used to estimate the advantages of different actions based on the rewards received and the estimated value function.
The policy update step in PPO involves computing the surrogate objective, which quantifies the improvement in the policy. PPO introduces a clip function that limits the policy update to a specified range, preventing large deviations from the previous policy. By constraining the policy updates, PPO ensures stability during training and mitigates the risk of policy divergence.

The methodology is based on the actor-critic approach, in which two neural networks interact with each other. In PPO, the actor is responsible for generating actions based on the current policy, while the critic estimates the value function and provides feedback on the quality of the chosen actions. By combining both models, the actor-critic approach enables the model to learn in an autonomous way from the critic's feedback to update the current policy, leading to a better decision-making actor.
%
%
%
%
Overall, the PPO algorithm provides a robust and stable framework for training reinforcement learning agents. Its conservative policy updates, combined with the use of surrogate objectives and entropy regularisation, enable efficient exploration and exploitation, leading to the discovery of optimal policies in complex environments, such as polynomial adaptation in high order methods.

The PPO agent has been trained through the Tensorforce Python library \cite{Tensorforce}, which provides a wide set of reinforcement learning algorithms to train the agent. Furthermore, it offers a flexible framework and the user has a complete control over the definitions of the environment, the state and the reward, as well as the value of the different hyperparameters involved in the process. Some of the most important hyperparameters are the learning rate, $lr = 10^{-3}$, the exploration factor, $\varepsilon = 10^{-2}$, the discount factor, $\gamma = 0.99$, and the likelihood ratio clipping, $rc = 0.1$.

The actor-critic approach used to train the agent requires two neural networks, both with the same architecture for this case. In addition to the input layer, with the shape of the state vector (see section \ref{state}), and the output layer, which provides a single output, each network has two hidden layers with 64 neurons each and an activation function of $\tanh$. The output of the actor network discriminates among the three possible actions through a linear function. Furthermore, the Adam optimiser \cite{kingma2014adam} has been used to train the networks. 

\subsubsection{State definition}\label{state}
We call state the set of variables that provides enough information to characterise  different situations. Based on this state, the agent can learn the best course of action for each scenario. In this case, we have defined a state based on two variables: the current polynomial degree, $p$, and an error, $e$. The polynomial degree will range as $p \in [2, 10]$, which is reasonable for solving PDEs with DGSEM. The error is a key point of the RL strategy and must provide an estimate of the accuracy that can be achieved when selecting a polynomial degree $p$.

\begin{figure}[h]
    \centering
    \includegraphics[width=0.5\textwidth]{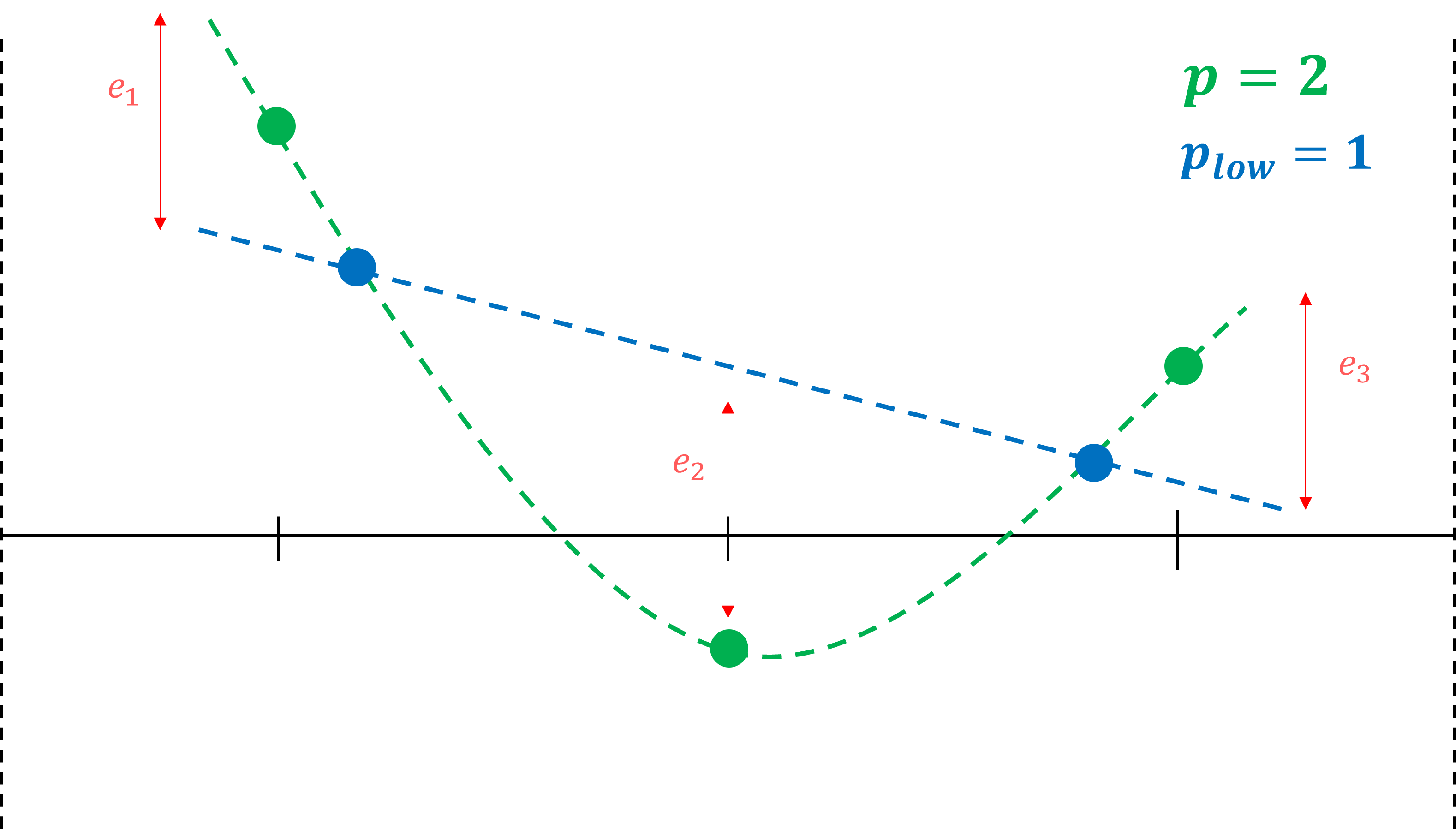}
    \caption{Error calculation between two consecutive polynomials of order $p$ and $p-1$.}
    \label{fig:state_error}
\end{figure}

When using a nodal high order approach (e.g., DGSEM in our case), we pose the equation at $p+1$ nodes (in our case Gauss nodes) in each element, given a polynomial of order $p$.
Having computed the numerical value at the nodes, it is possible to compute/reconstruct the continuous solution inside each element using Lagrange polynomials. Then, the polynomial can be used to interpolate the solution for the nodes associated with a lower degree $p-1$. The low-degree solution is then compared with the previous solution to compute $N_s = 3$ errors, $e_i$ for $i=1, 2 \; \mathrm{and} \; 3$, which are obtained by subtracting the low-order solution from the high-order solution in the middle and at the boundaries of the element, as shown in Figure~\ref{fig:state_error}. This approach is selected because it is fast and considers the boundary values of the elements. This is interesting because the inter-element jumps are related to the error in the solution in DG methods \cite{ferrer2017interior,kou2022jump}.
If the error between two consecutive polynomials is small, then the shape of the true solution is accurately captured, and hence that polynomial order is high enough to obtain a precise numerical approximation. Finally, these three errors are combined to obtain a single scalar value $mse$:
\begin{eqnarray}
    mse = \frac{\sum_{i=1}^{N_s} {e_i}^2}{N_s},\\
    \label{eq:state_rmse}
    e = -\log_{10}\left( \min \left(mse + 10^{-10}, 1 \right) \right).
    \label{eq:state_error}
\end{eqnarray}

The error, $e$, has been mapped using the logarithm to separate values of different orders of magnitude. The image of the error is $e \in [0, 10]$. In this way, the values of $e$ and $p$, have similar ranges, which is advantageous for the training process. 
We favour this estimation of the error because of its efficiency (only 3 points need to be evaluated) and its locality (all information is contained in the element), but of course other error estimations could be used (e.g., \cite{doi:10.2514/6.2006-112,rueda2019p,laskowski2022functional}).
As the error, $e$, is a continuous variable, an infinite number of states $(p, e)$ are possible; and hence, the selected RL model must be capable of handling continuous states, as is the case for the PPO model which is used in this work.

\subsubsection{Reward definition}\label{reward}
We define the reward as the objective function that the agent should learn to decide the best course of action based on the current state. This function should have a higher value as the error decreases. In addition,  it should have a higher value as the polynomial order decreases, because the objective is to use the minimum polynomial order possible to reduce the computational time while preserving the accuracy.

\begin{figure}[h]
\centering
  \begin{subfigure}[b]{0.48\textwidth}
    \includegraphics[width=\textwidth]{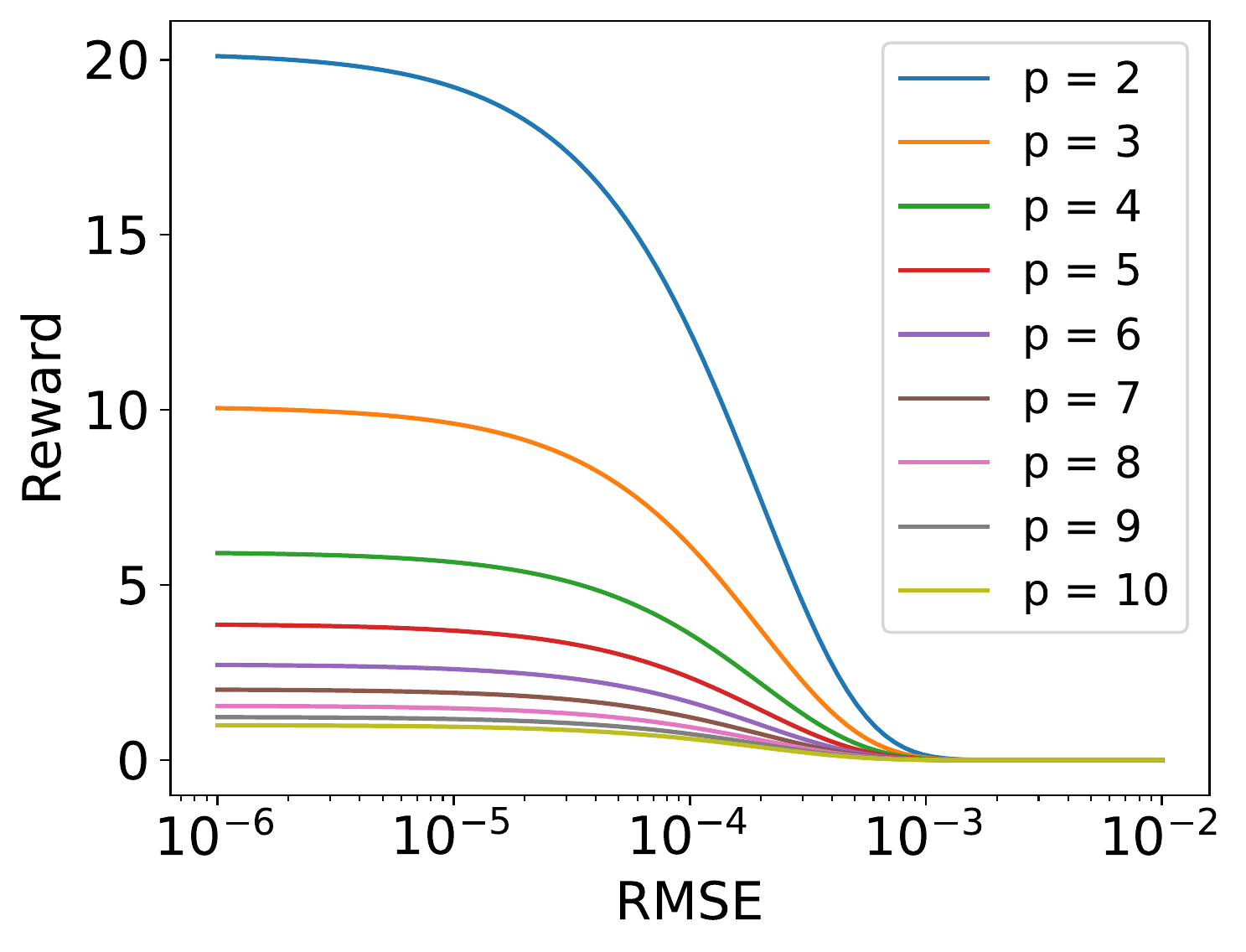}
    \caption{Reward vs $rmse$}
    \label{fig:reward_rmse}
  \end{subfigure}
  \quad
  \begin{subfigure}[b]{0.48\textwidth}
    \includegraphics[width=\textwidth]{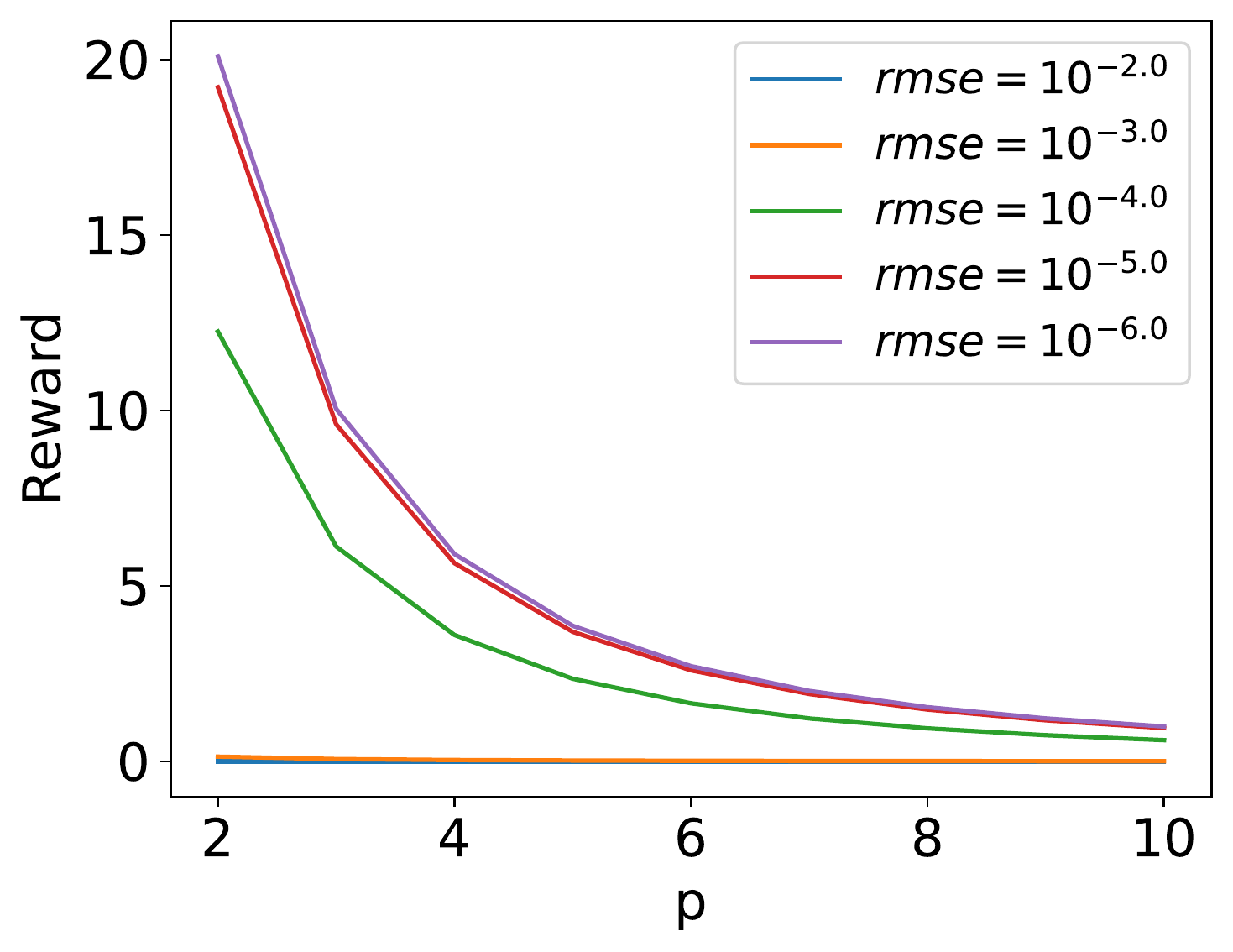}
    \caption{Reward vs p}
    \label{fig:reward_p}
  \end{subfigure}
  
\caption{Reward distribution as a function of a) the error $rmse$ and b) the polynomial order $p$, both with a value of  $\sigma = 0.01$.}
\label{fig:reward_function}
\end{figure}

\begin{figure}[h]
    \centering
    \includegraphics[width=0.48\textwidth]{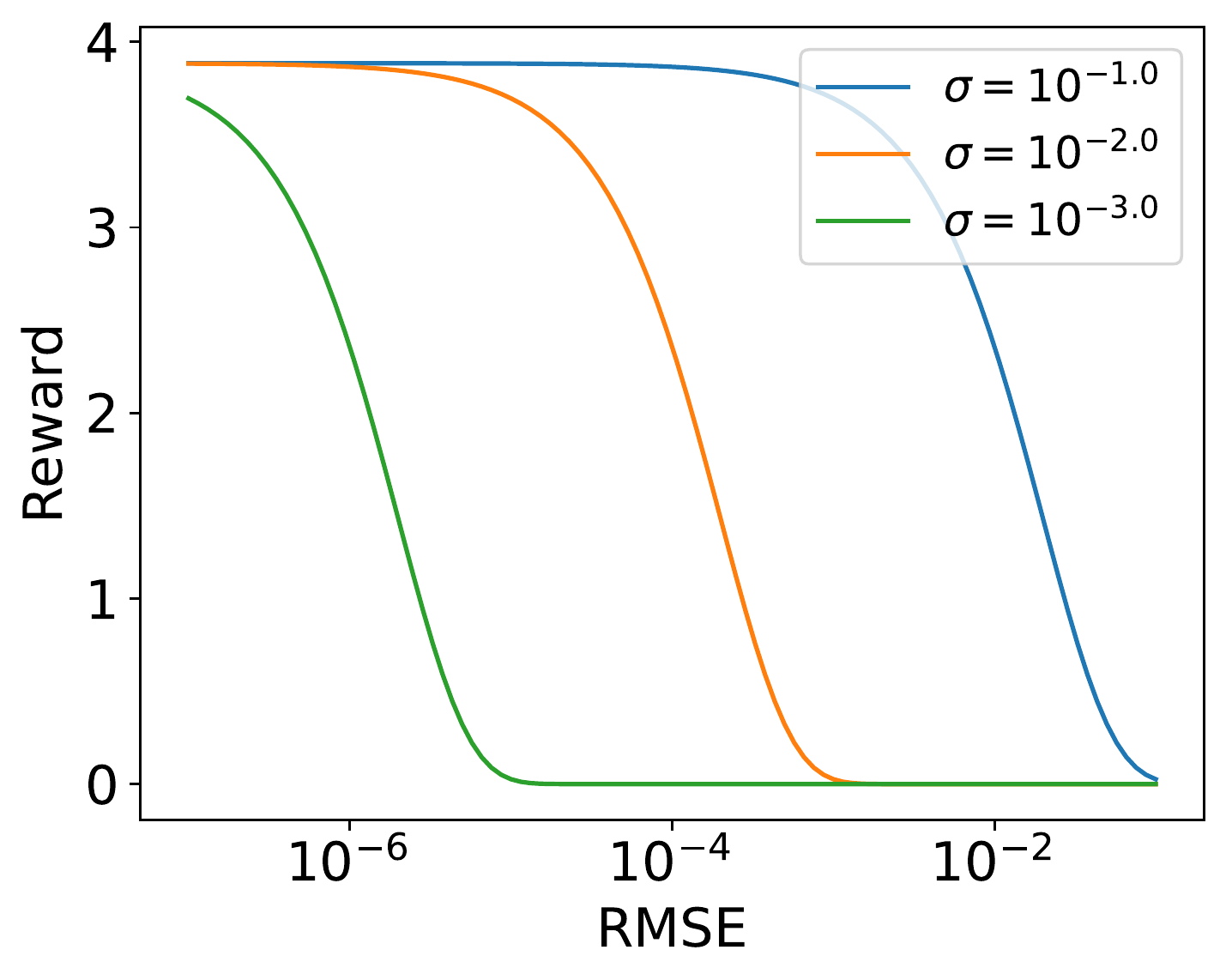}
    \caption{Reward distribution as a function of the error $rmse$ for three different values of $\sigma$ and a constant polynomial order $p=5$.}
    \label{fig:reward_sigma}
\end{figure}

The error used to calculate the reward should not be the same as that used for the state, as the reward is only computed during the training phase and it must provide a measurement of the real accuracy of the model. In this case, we require the error between our polynomial approximation and the analytical solution, which is a reliable value for evaluating the performance of the model. Therefore, the analytical solution (or an estimation) must be known during the training, as will be explained later in section \ref{training}. We use the root mean squared error, $rmse$, as the difference between the analytical and approximate ($p-1$) solutions sampled at $N_r$ Gauss points. In this work, we use a constant value of $N_r = 2p_{max}+1$ which is enough to accurately approximate Eq. \eqref{eq:rmse_reward}. In our proposed strategy, we calculate an approximate solution using the polynomial order $p-1$ instead of $p$. By doing so, we reward the accuracy of the solution $p-1$, which is the lowest order required to compute the state (see the previous section \ref{state}), leading to a more robust method.
We finally, define the reward function as a smooth Gaussian-like distribution, which provides a trade-off between cost (low polynomial order) and accuracy (low error):
\begin{equation}
    reward = \overbrace{\frac{p_{\max}^2+1}{p^2+1}}^\text{COST} \overbrace{\exp\left({-\frac{rmse^2}{2 \sigma^2}}\right)}^\text{ERROR},
    \label{eq:reward}
\end{equation}
\begin{equation}
    rmse = \sqrt{\frac{\sum_{i=1}^{N_r} (y_i - y^*_i)^2}{N_r}},
    \label{eq:rmse_reward}
\end{equation}
with $p_{\max}=10$ as the maximum polynomial order available , $\sigma$ as the standard deviation, $y$ as the analytical solution and $y^*$ as the polynomial approximation. 
 The main contribution of the reward is the exponential part (the error), which will drop the reward if the error is big compared to the standard deviation. The second contribution is the quadratic part $\frac{p_{\max}^2+1}{p^2+1} $, it relates to the computational cost, and only becomes relevant if the error $rmse$ is small enough. The cost part increments the reward if the polynomial order is small. Consequently, the reward enforces the error to be small and, while keeping the polynomial degree as small as possible (but without increasing the error significantly). We illustrate the behaviour of the reward as a function of $rmse$, $p$ and $\sigma$ in Figures~\ref{fig:reward_function} and \ref{fig:reward_sigma}. Figure~\ref{fig:reward_rmse} shows the exponential decay of the reward (the error part) with the $rmse$ and Figure~\ref{fig:reward_p} shows the quadratic part (the cost) of the reward with the polynomial order. The value of $\sigma$ is important, as it provides an estimation of the value of $rmse$ that must be achieved before the polynomial order can be optimised. If the value of $\sigma$ is small, the agent will try to significantly reduce the $rmse$, which will lead to an adaptation with high polynomials, and vice versa. The effect of $\sigma$ on the reward is shown in Figure~\ref{fig:reward_sigma}. The final error $rmse$ (after the adaptation process) will have an approximate value defined by the region where the reward is increasing with its maximum slope (in Figure~\ref{fig:reward_sigma}). 
 This way, an agent can be trained to achieve a specific threshold error, while minimising the cost. Based on preliminary tests (not shown), we select $\sigma = 0.01$, which provides an appropriate trade-off between accuracy and computational cost of the solution. This is what defines the efficiency and accuracy of the p-adaptation strategy.

\newpage

\subsubsection{Training}\label{training}
During the training process, the agent interacts with the environment and learns which action provides the highest reward for each state. In our training, the agent learns to modify the polynomial order of one element of the mesh. Within the DGSEM framework, every element of the mesh is mapped into the computational space where $x \rightarrow \xi \in [-1,1]$, see \ref{app:dgsem} for details. We can simplify and optimise the training process if the agent is trained to optimise the polynomial order in computational space. This approach allows to use a common agent for every element of the mesh.

One of the main problems of using RL for numerical applications is that the training is very time consuming. Furthermore, the reward, Eq. \eqref{eq:reward}, is based on an error that is computed in relation to the analytical solution, which is not known for most partial differential equations (e.g., Navier-Stokes equations). 
To tackle both problems, in this work we propose to decouple the training process of the RL agent from the simulation. The training is performed using a random function generator, which provides random sinusoidal-type functions in one element. 
This random function generator can create functions, $g:[-1,1]\rightarrow \mathbb R$, with the following structure:
\begin{equation}
    g(\xi) = \frac{1 + a \cdot \sin(2  \pi  f \, (\xi - c))}{2}
    \label{eq:random_function_generator}
\end{equation}
where $a = \{-1,1\}$, $c$ $\in [-1,1]$ are random parameters and $f\in [0,1]$ is a random frequency. More complex functions can be created to capture specific behaviours, but this simple option shows promising results. 
Let us note that when correlating polynomials and sinusoidal functions, we are, in fact, performing a non-linear von Neumann (or eigensolution analysis) type analysis \cite{manzanero2018dispersion,kou2021VonNeumann,KOU2022110798} and consequently we are finding the optimal polynomials that capture wavelike solutions.\\ 

The training process is divided in episodes, each one formed by the following steps:
\begin{enumerate}
    \item A random function is generated using equation \eqref{eq:random_function_generator} and the initial polynomial order $p \in [2,10]$ is chosen randomly.
    \item The initial state is computed.
    \item The agent is called to select an action based on the current state.
    \item The action is performed, and a new state and a reward are generated.
    \item The agent learns whether the chosen action was good or not based on the reward.
    \item Steps 3) to 5) are repeated $N = 20$ times.
\end{enumerate}
The process is repeated $N_{ep} = 25000$ episodes until the model is trained.\\ 

\begin{figure}[h]
\centering
  \begin{subfigure}[b]{0.48\textwidth}
    \includegraphics[width=\textwidth]{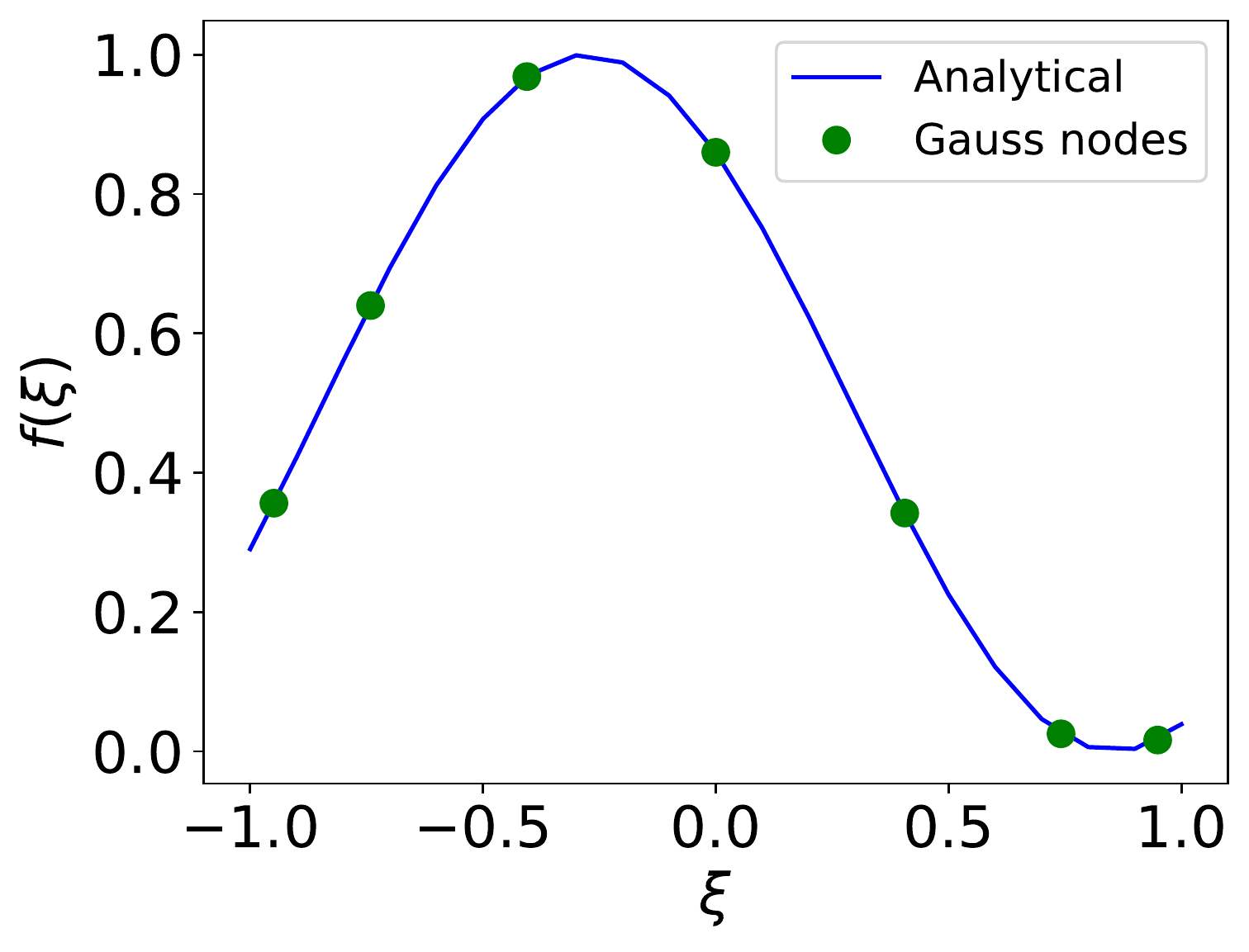}
    \caption{$p=6$}
    \label{fig:random_function_p6}
  \end{subfigure}
  \quad
  \begin{subfigure}[b]{0.48\textwidth}
    \includegraphics[width=\textwidth]{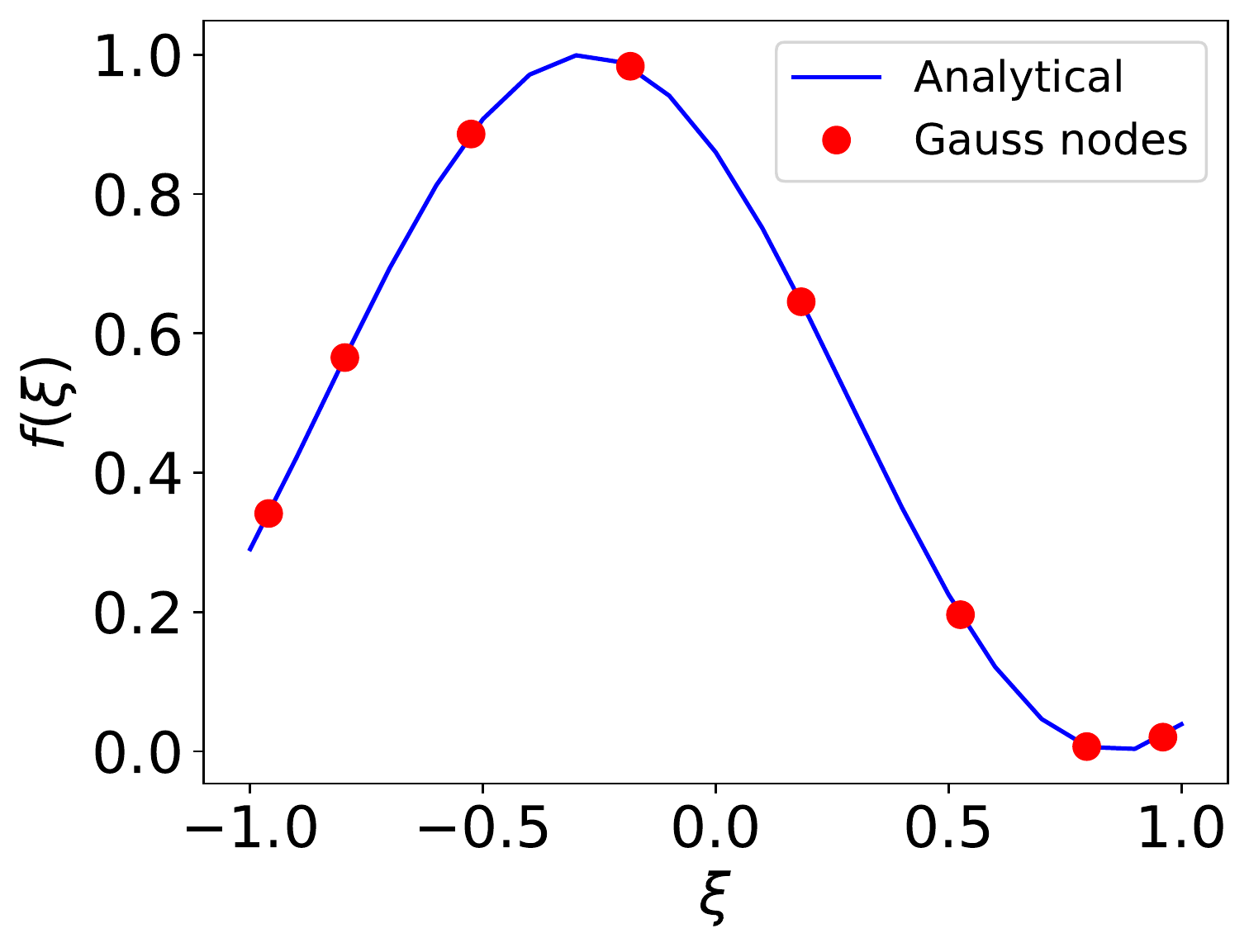}
    \caption{$p=7$}
    \label{fig:random_function_p7}
  \end{subfigure}
  
    \centering
    \caption{Random function generated during the training and the Gauss nodes a) for the initial value of $p=6$ and b) for its value after the action is executed $p_{new}=7$.}
    \label{fig:random_function}
\end{figure}

Let us clarify the methodology using an example.
First, a initial polynomial order, $p$, is selected randomly ($p=6$ in this example) and Eq. \eqref{eq:random_function_generator} is used to generate a random analytical sinusoidal function. This analytical solution is sampled at the Gauss nodes corresponding to the selected polynomial, as represented in Figure~\ref{fig:random_function_p6}. Then, the value at the nodes is used to construct a Lagrange interpolating polynomial, which approximates the solution. Knowing the approximate sampled solution (for $p=6$ in this example), the initial state can be calculated, following section \ref{state}.

The initial state is the input for our agent that provides the course of action. As explained above, the agent can choose among three possible actions: to increase the polynomial order by one ($p=7$), to decrease the polynomial order by one ($p=5$) or to keep the current polynomial order ($p=6$). For this example, let us consider that the action decided by the agent is to increase the polynomial order; that is, to change the current $p=6$ to $p_{new}=7$.

Given this new polynomial, the value of the solution at the Gauss nodes is calculated from the analytical solution (Eq. \eqref{eq:random_function_generator}), which is the same function as before. The new sampled Gauss nodes are represented in Figure~\ref{fig:random_function_p7}.
These new values are then used to generate a new Lagrange polynomial to approximate the solution. For this condition, a new state (see section \ref{state}) and a reward (see section \ref{reward}) are computed. 

Finally, the agent stores the initial state, the action chosen and the reward to update its policy at the end of the episode. This update is the core of the training process, and it allows the agent to learn the best course of action for a given state.

The previous steps are repeated several times (we have selected $N=20$ times per episode). 
After $N$ iterations, the episode finishes, the agent updates the internal weights of the networks, and a new episode begins with a new random function as the analytical solution.
  
\subsubsection{Testing}\label{testing}
\begin{figure}[h]
    \centering
    \includegraphics[width=0.5\textwidth]{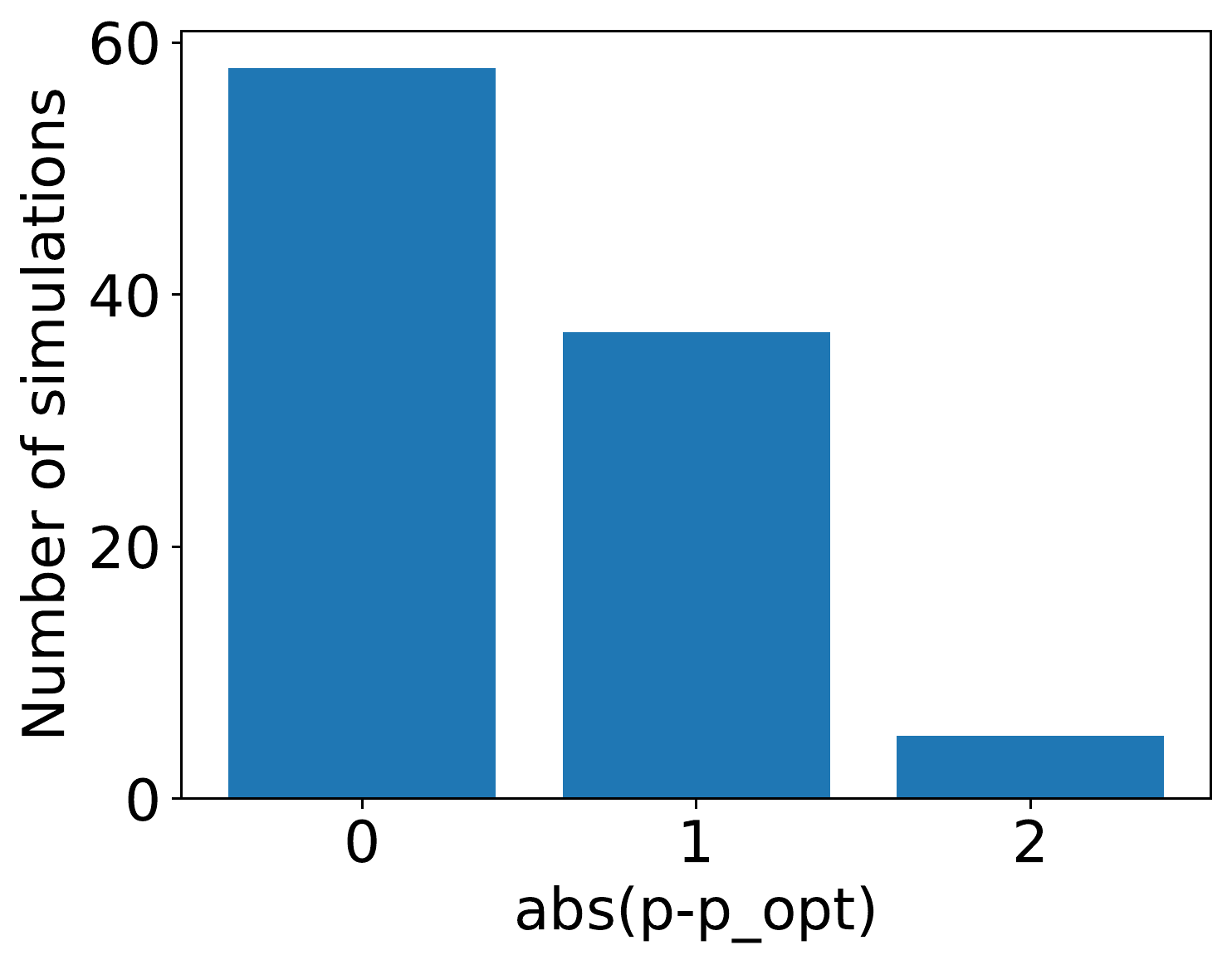}
    \caption{Absolute error between the optimum polynomial order (based on the reward function) and the one selected by the RL agent in 100 simulations.}
    \label{fig:histogram_error_p}
\end{figure}

Once the model has been trained, the agent can be tested by generating a new set of random functions and comparing the maximum reward obtained for each case using RL with the expected optimal reward. Each test set is formed by 100 simulations of random functions, and the accuracy of the rewards and the mean error of the polynomial order for the RL agent are computed. The results are summarised in Table~\ref{tab:test_results}. 
In each test set, the accuracy obtained by the model is close to $90\%$; that is, the mean reward obtained in the simulations is $90\%$ of the maximum possible value, which represents very good performance. 
The mean error between the optimal polynomial order and the one selected by the agent is close to 0.5, which means that most of the simulations ($50\%$ at least) have been perfectly performed by the agent. This result is represented in Figure~\ref{fig:histogram_error_p}, which shows the
absolute error between the optimum polynomial order (based on the reward function) and the one selected by the RL agent in 100 simulations. It can be seen that the agent can select the best polynomial order the $60\%$ of the time. Of course, the agent sometimes selects one degree less than or more from the optimum. 
In general, the trained model is accurate and capable of performing a competitive p-adaptation over a wide set of functions, and it can be used to improve the performance of numerical simulations.

\begin{table}[h]
    \centering
    \begin{tabular}{|c|c|c|}
    \hline
    \textbf{Test set} & \textbf{Reward accuracy} & \textbf{Mean error $|p-p_{opt}|$}  \\ \hline
    
    1 & 0.875 & 0.48 \\ \hline
    2 & 0.864 & 0.59 \\ \hline
    3 & 0.875 & 0.56 \\ \hline
    4 & 0.862 & 0.53 \\ \hline
    5 & 0.889 & 0.49 \\ \hline
    6 & 0.874 & 0.54 \\ \hline
    7 & 0.880 & 0.51 \\ \hline
    8 & 0.869 & 0.55 \\ \hline
    9 & 0.848 & 0.59 \\ \hline
    10 & 0.864 & 0.49 \\ \hline
    
    \end{tabular}
    \caption{Accuracy of the rewards in relation to the maximum possible value and mean error of the polynomial order for the RL agent compared with the optimal polynomial based on the reward function. Each test set is formed by 100 simulations of random functions.}
    \label{tab:test_results}
\end{table}

\newpage
\section{Results}\label{res}
To demonstrate the effectiveness of the proposed methodology, the trained RL agent is used to improve the accuracy of a Burgers' equation, Eq. \eqref{eq:Burgers_equation}. In a separate phase, the PPO agent has learnt how to increase or decrease the polynomial order of each element of a given mesh to obtain an optimum performance; that is, a very low error at the lowest computational cost.

\begin{figure}[h]
    \centering
    \includegraphics[width=0.5\textwidth]{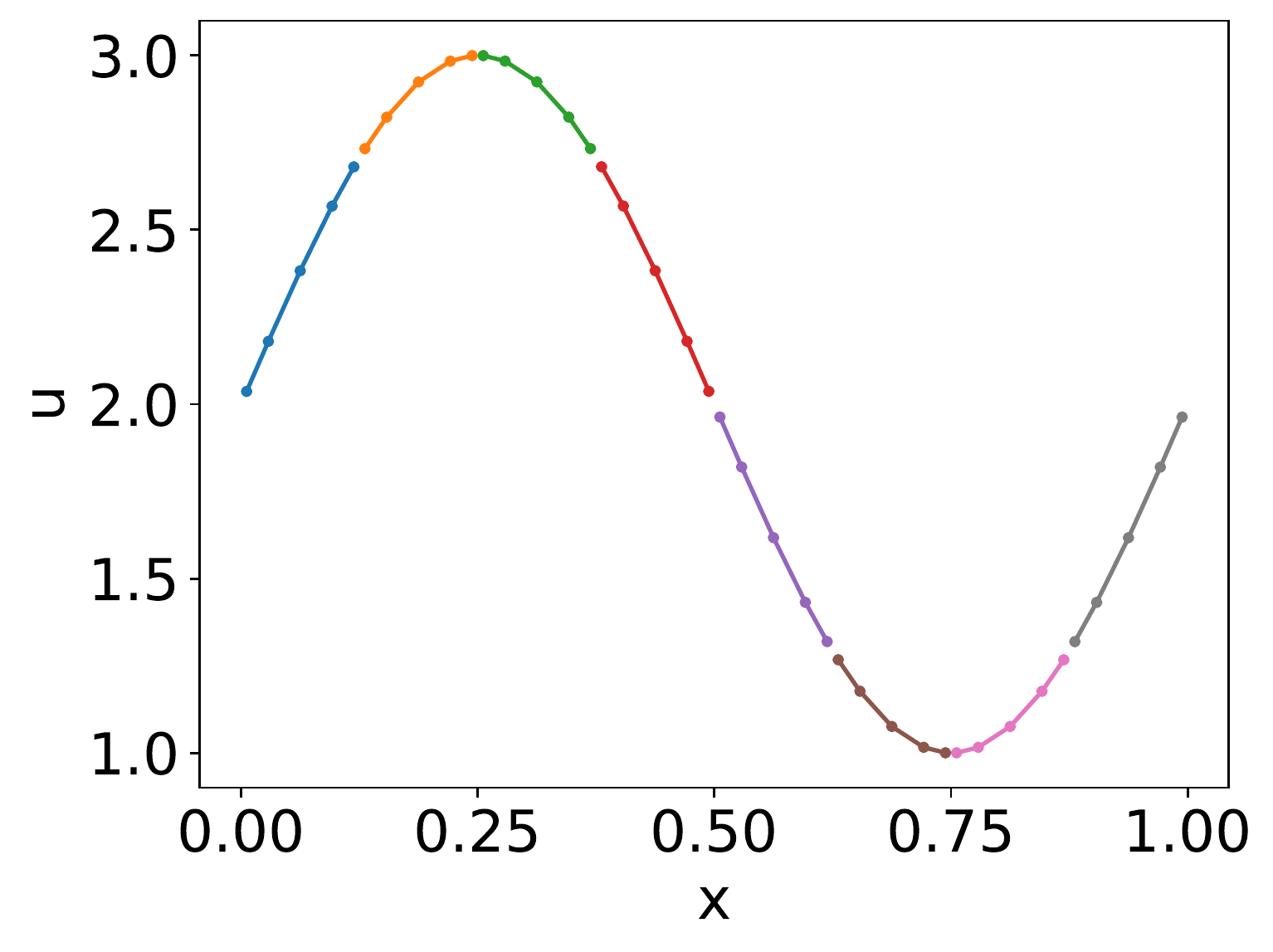}
    \caption{Initial condition to solve the 1D inviscid Burgers' equation with periodic BC.}
    \label{fig:Burgers_initial_condition}
\end{figure}

The computational mesh is made up of 8 elements and the initial polynomial degree is $p=4$ for every element, as shown in Figure~\ref{fig:Burgers_initial_condition}, where the initial solution is shown at Gauss points. A time step of $\Delta t = 10^{-5} \, \mathrm{s}$ is used to perform the temporal integration.

\begin{figure}[htbp]
  \centering
  \begin{subfigure}[b]{0.48\textwidth}
    \includegraphics[width=\textwidth]{No_adaptation_p4_t0.0000.pdf}
    \caption{$t=0.0 \, \mathrm{s}$}
    \label{fig:no_adaptation_t0}
  \end{subfigure}
  \quad
  \begin{subfigure}[b]{0.48\textwidth}
    \includegraphics[width=\textwidth]{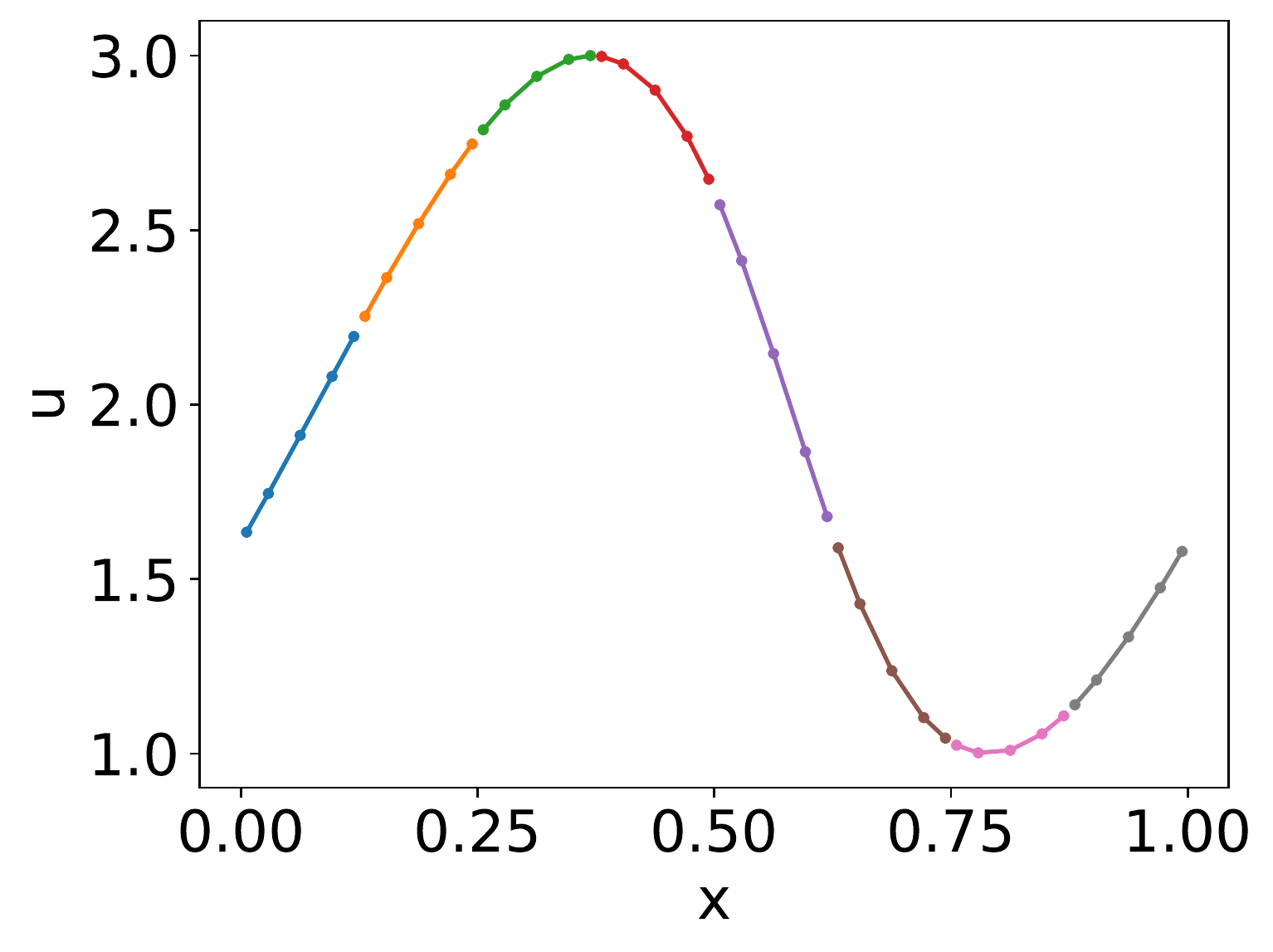}
    \caption{$t=0.04 \, \mathrm{s}$}
    \label{fig:no_adaptation_t0.04}
  \end{subfigure}
  
  \vspace{1cm} 
  
  \begin{subfigure}[b]{0.48\textwidth}
    \includegraphics[width=\textwidth]{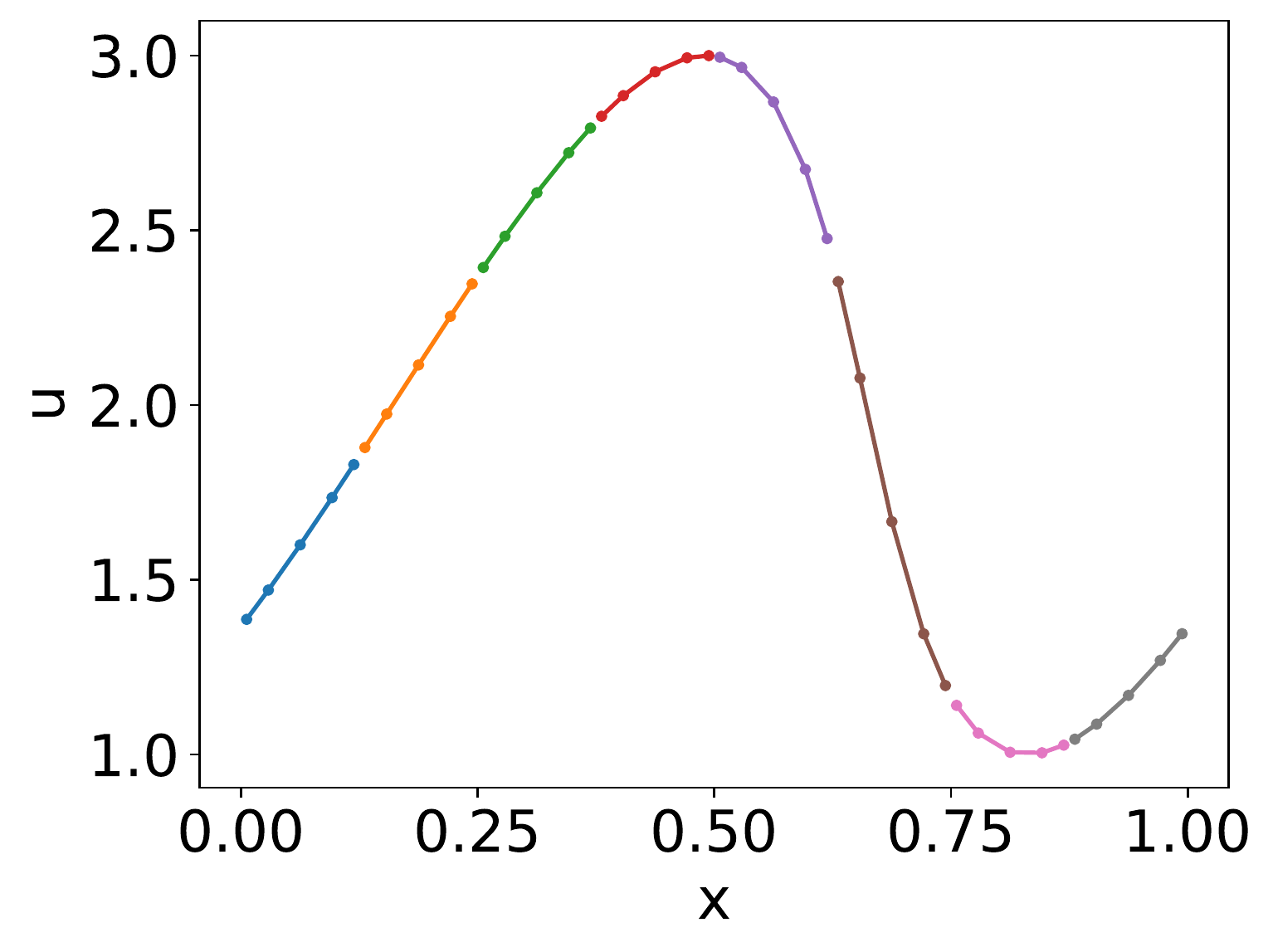}
    \caption{$t=0.08 \, \mathrm{s}$}
    \label{fig:no_adaptation_t0.08}
  \end{subfigure}
  \quad
  \begin{subfigure}[b]{0.48\textwidth}
    \includegraphics[width=\textwidth]{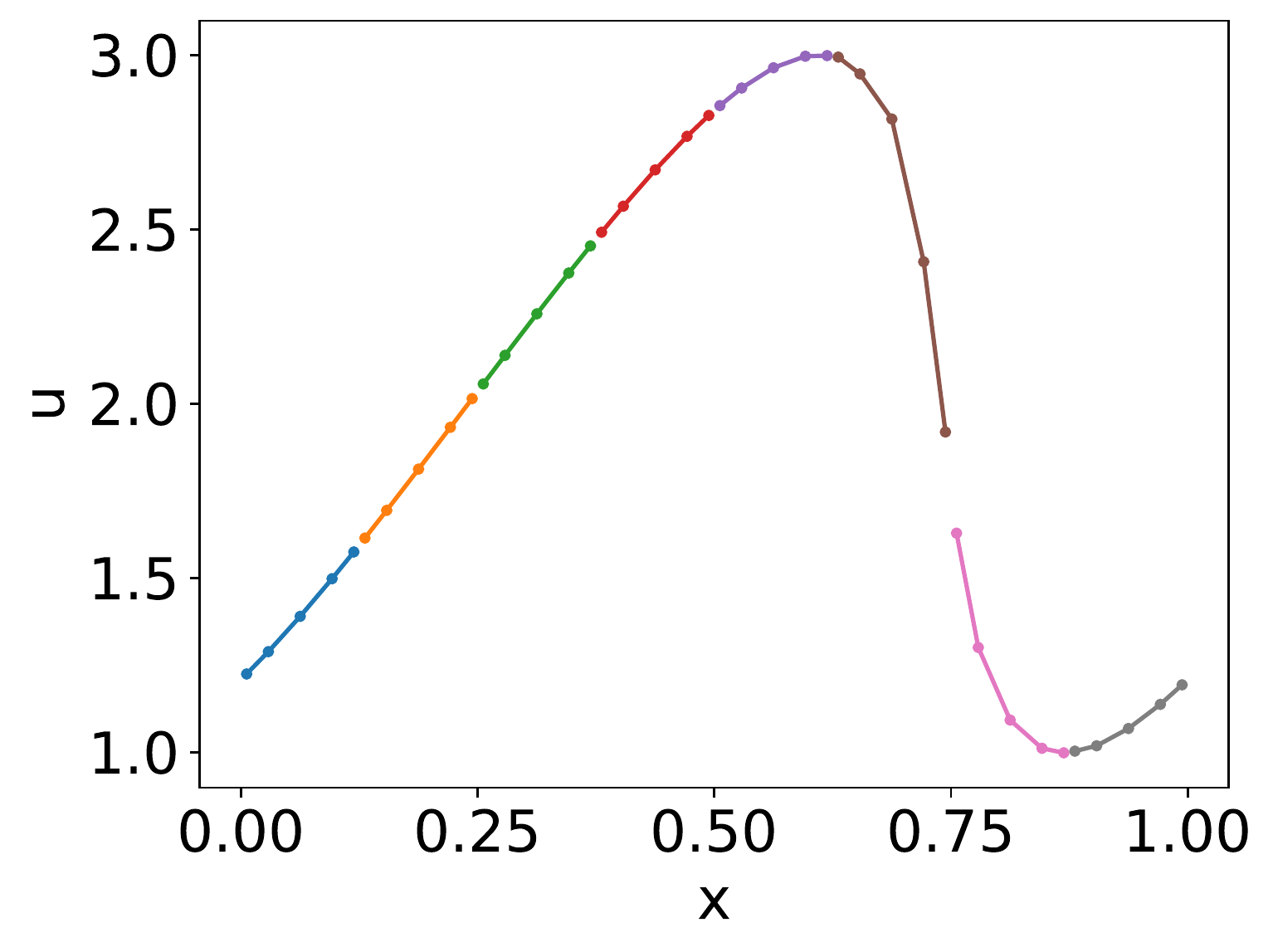}
    \caption{$t=0.12 \, \mathrm{s}$}
    \label{fig:no_adaptation_t0.12}
  \end{subfigure}
  
  \caption{1D inviscid Burgers' equation solved with a DGSEM spatial scheme and an explicit Euler temporal scheme, using a constant polynomial order, $p=4$, in each element of the computational mesh. The solution is shown in four different timestamps.}
  \label{fig:Burgers_no_adaptation}
\end{figure}

The solution of the problem without p-adaptation (uniform $p=4$) for $t \in [0, 0.12] \, \mathrm{s}$ is represented in Figure~\ref{fig:Burgers_no_adaptation}, where we have used colours to represent the different elements. Note that we stop the simulation before the shock forms, since we are interested in the smooth phase of this problem. Two main regions can be highlighted in the solution at the final time. On the one hand, we observe a left region, where the solution is almost linear, and the right region, where a strong gradient is present. The left region could be captured with a polynomial order $p=2$, because the velocity changes progressively and its gradient is almost constant. On the other hand, the right region requires a higher polynomial degree to capture the abrupt change of the solution. Therefore, the use of a constant polynomial order $p=4$ is far from ideal, as some elements use a polynomial order higher than required, which increases the computational cost, while other elements use a polynomial order lower than required, which deteriorates the precision of the solution. Both problems can be solved using the proposed RL strategy.

\begin{figure}[htbp]
  \centering
  \begin{subfigure}[b]{0.48\textwidth}
    \includegraphics[width=\textwidth]{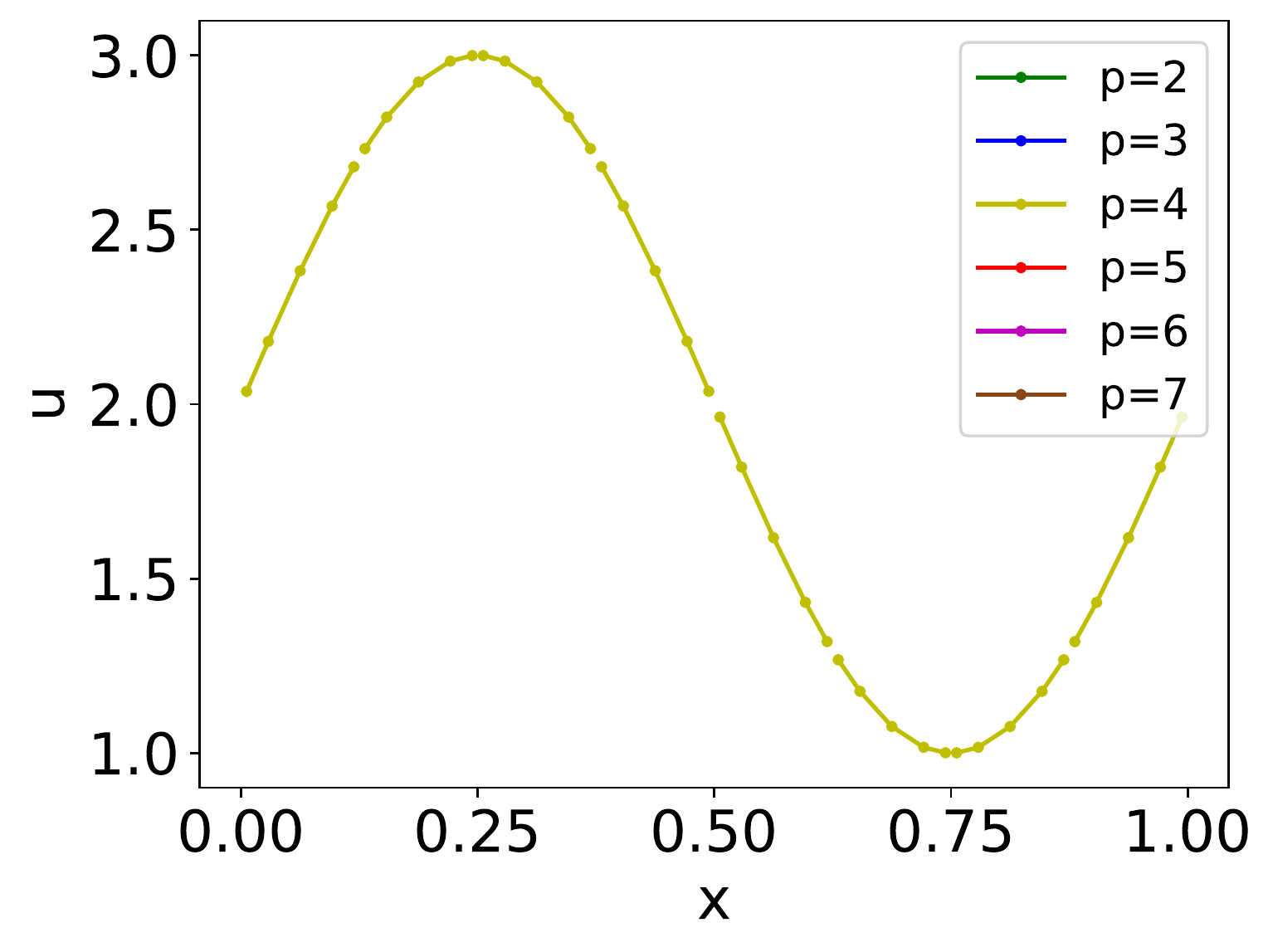}
    \caption{$t=0.0 \, \mathrm{s}$}
    \label{fig:adaptation_t0}
  \end{subfigure}
  \quad
  \begin{subfigure}[b]{0.48\textwidth}
    \includegraphics[width=\textwidth]{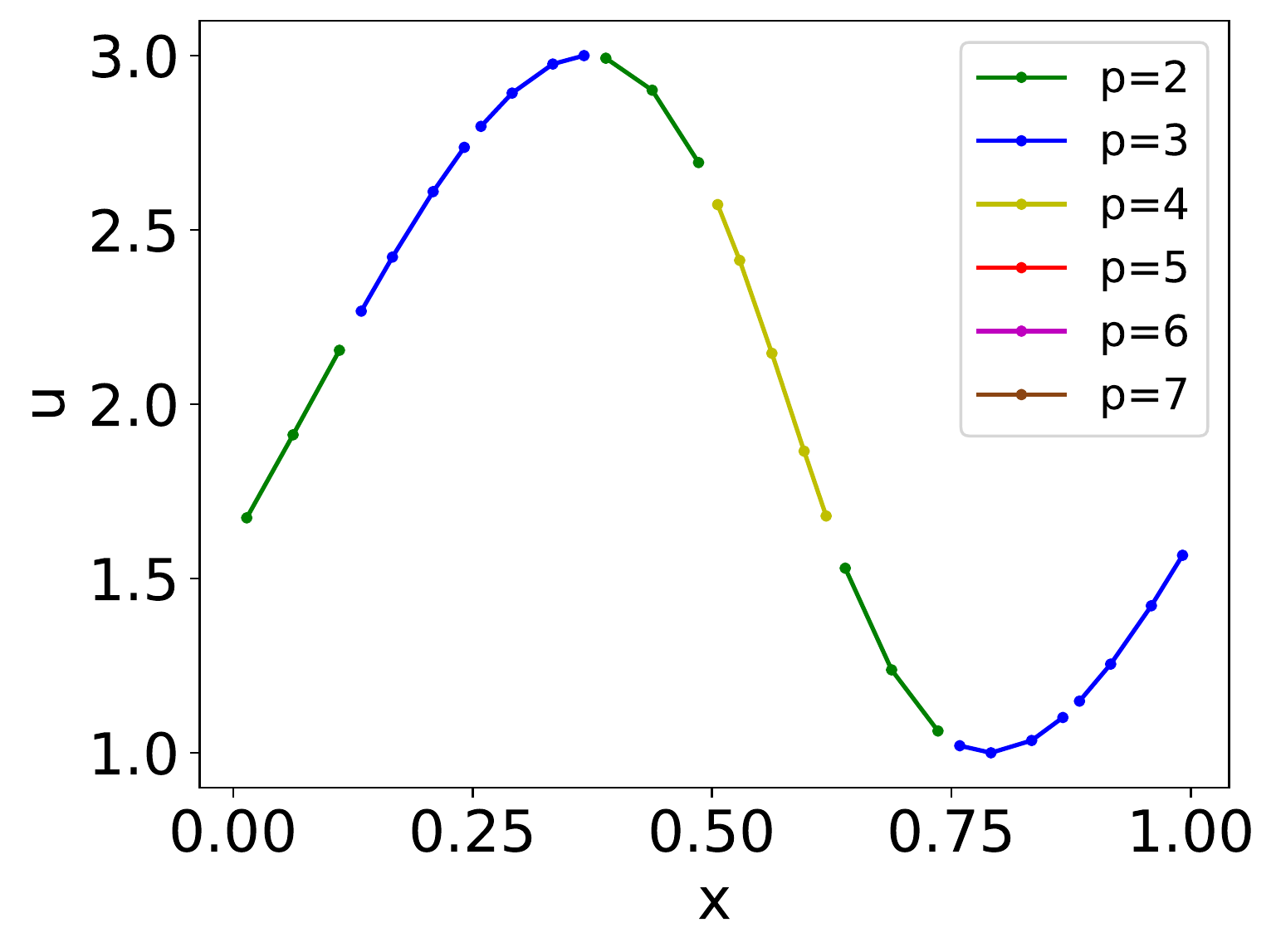}
    \caption{$t=0.04 \, \mathrm{s}$}
    \label{fig:adaptation_t0.04}
  \end{subfigure}
  \vspace{1cm} 
  \begin{subfigure}[b]{0.48\textwidth}
    \includegraphics[width=\textwidth]{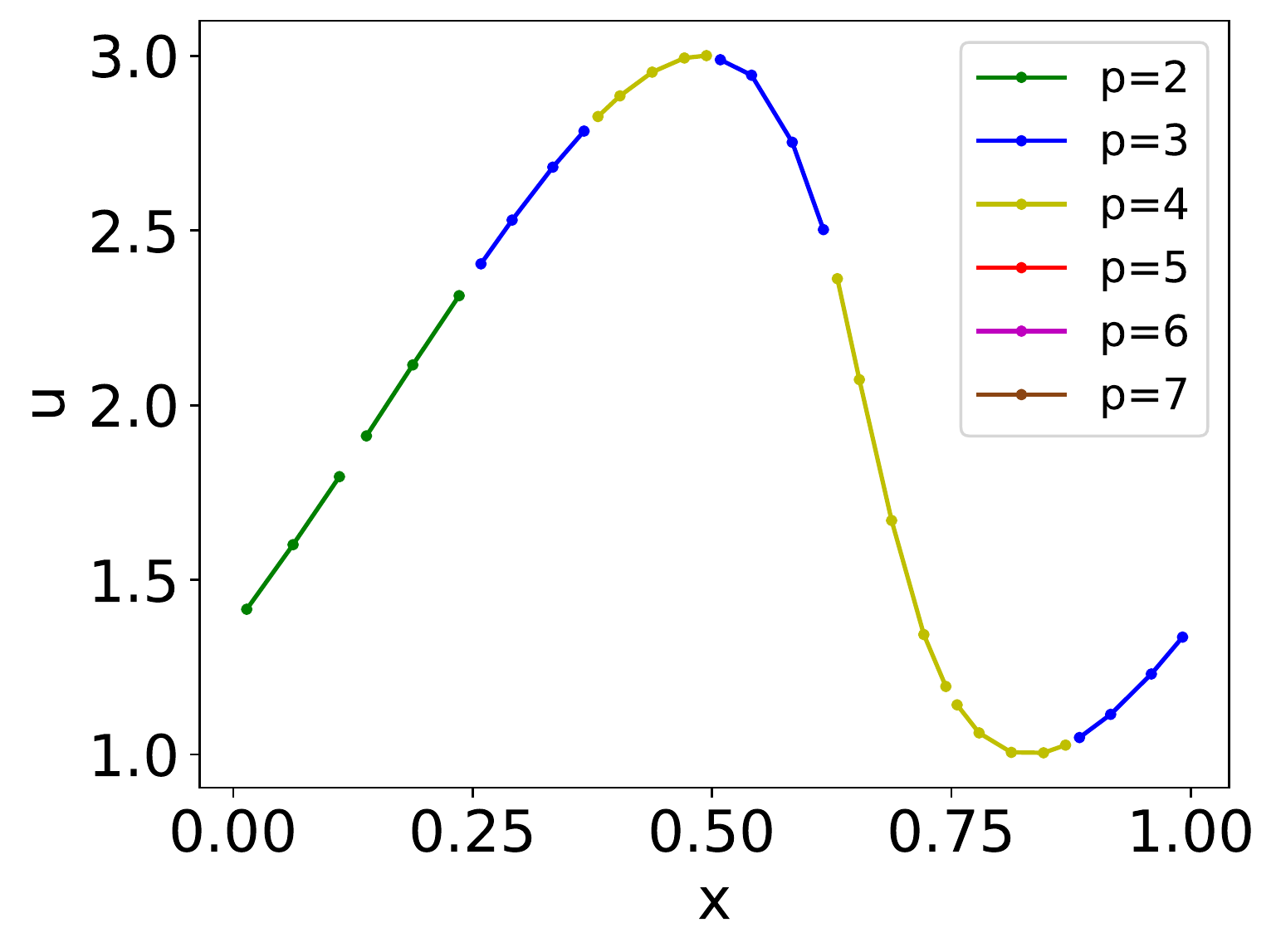}
    \caption{$t=0.08 \, \mathrm{s}$}
    \label{fig:adaptation_t0.08}
  \end{subfigure}
  \quad
  \begin{subfigure}[b]{0.48\textwidth}
    \includegraphics[width=\textwidth]{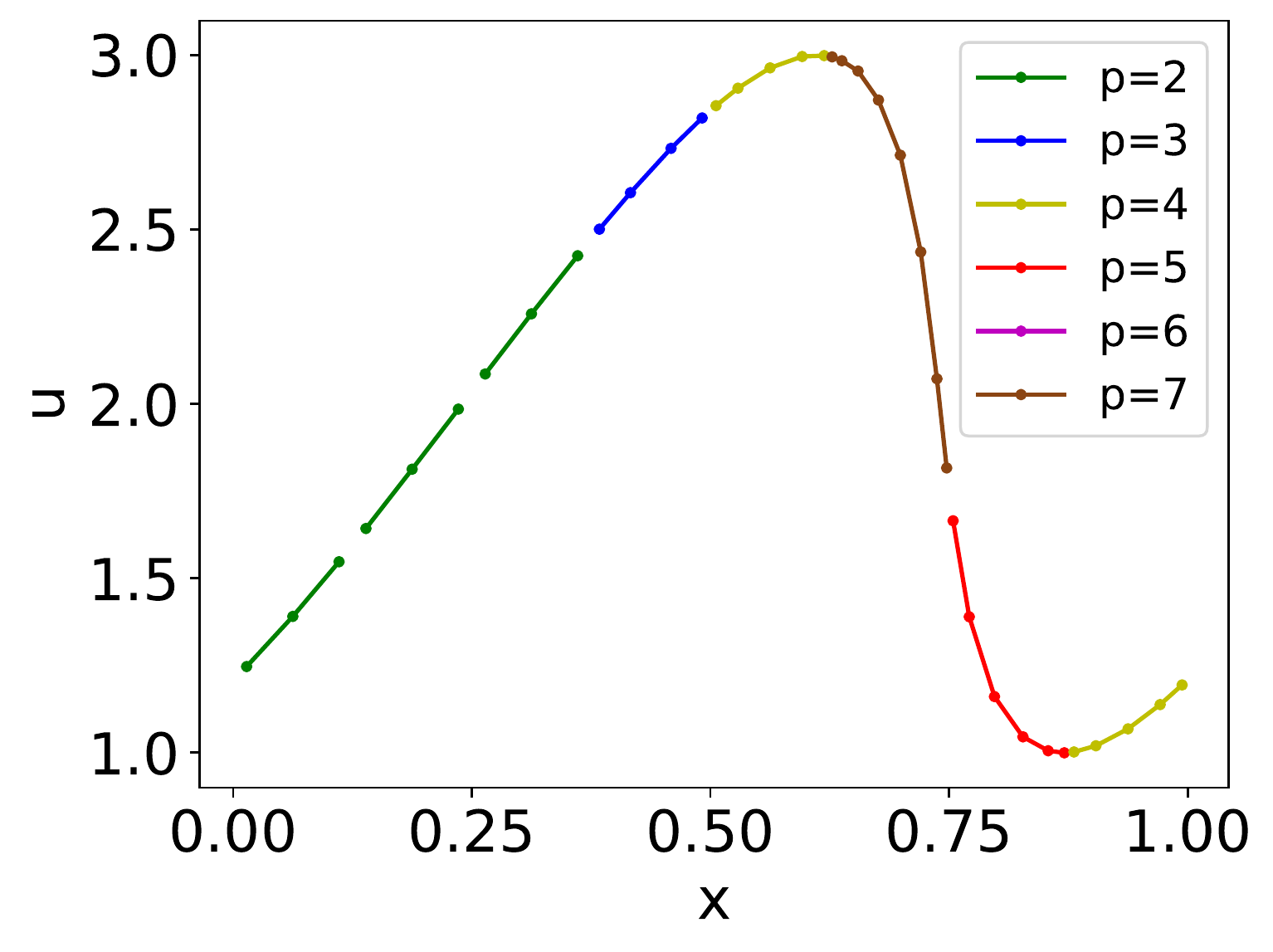}
    \caption{$t=0.12 \, \mathrm{s}$}
    \label{fig:adaptation_t0.12}
  \end{subfigure}
  
  \caption{1D inviscid Burgers' equation solved with a DGSEM spatial scheme and an explicit Euler temporal scheme, using a RL p-adaptation algorithm. The polynomial of each element of the computational mesh has been adapted once every $0.01 \, \mathrm{s}$ of simulation time, starting with a polynomial order $p=4$ in every element. The solution is shown in four different timestamps.}
  \label{fig:Burgers_adaptation}
\end{figure}

We now simulate the same problem using the RL p-adaptation strategy. The solution for $t \in [0, 0.12] \, \mathrm{s}$ is represented in Figure~\ref{fig:Burgers_adaptation}. The adaptation process is performed each $0.01 \, \mathrm{s}$ of simulation time, that is each 1000 iterations, for every element in the mesh. To improve the adaptation process, the solution within each element has been normalised between 0 and 1 before the computation of the state. In this way, the input will be more similar to the states used during the training.

In this case, the RL model is able to dynamically modify the polynomial order, reducing the order when possible and increasing the order near regions with strong gradients.
When the agent decides to modify the polynomial order inside one element, the interpolated solution is used to continue the simulation.
The final solution is more accurate and less time consuming than the previous solution without p-adaptation (computational costs are compared in section \ref{tradeoff}).

\begin{figure}[htbp]
  \centering
  \begin{subfigure}[b]{0.4\textwidth}
    \includegraphics[width=\textwidth]{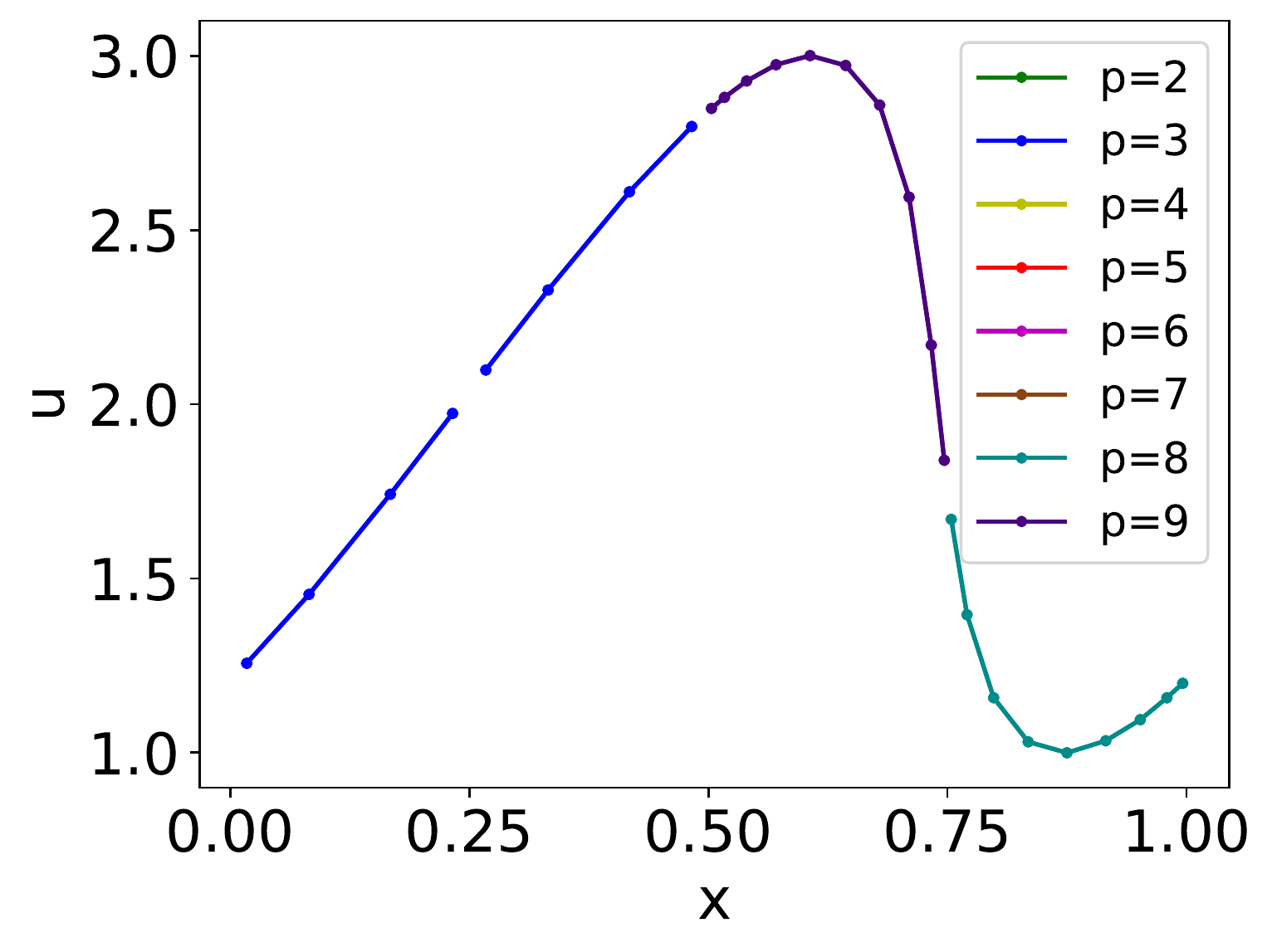}
    \caption{4 elements}
    \label{fig:adaptation_t0.12_elements4}
  \end{subfigure}
\quad
  \begin{subfigure}[b]{0.4\textwidth}
    \includegraphics[width=\textwidth]{Adaptation_p4_t0.1200.pdf}
    \caption{8 elements}
    \label{fig:adaptation_t0.12_elements8}
  \end{subfigure}
%
  %
  \begin{subfigure}[b]{0.4\textwidth}
    \includegraphics[width=\textwidth]{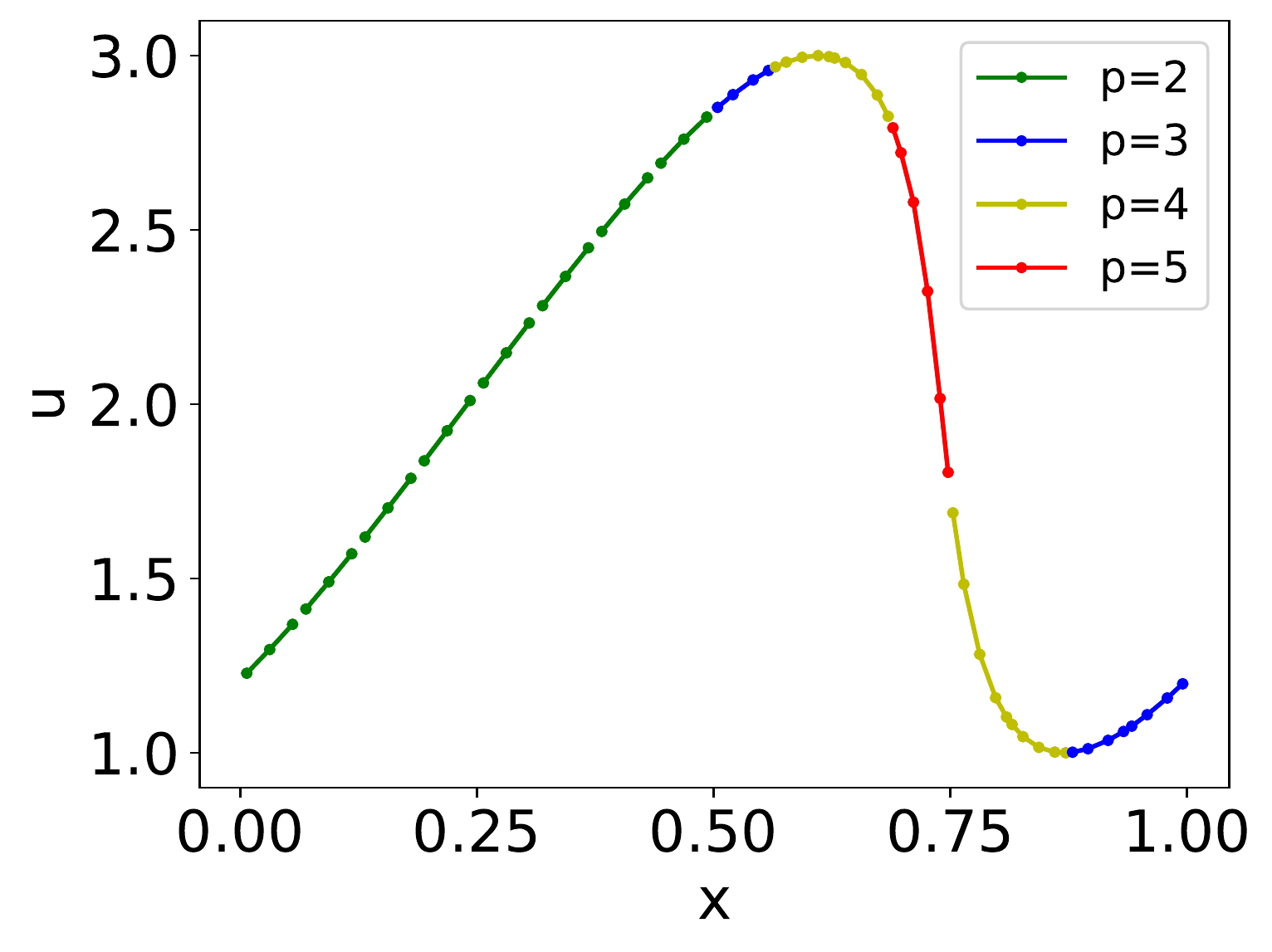}
    \caption{16 elements}
    \label{fig:adaptation_t0.12_elements16}
  \end{subfigure}
  
  \caption{1D inviscid Burgers' equation solved with a DGSEM spatial scheme and an explicit Euler temporal scheme, using a RL p-adaptation algorithm. The polynomial of each element of the computational mesh has been adapted once every $0.01 \, \mathrm{s}$ of simulation time, starting with a polynomial order $p=4$ in every element. The solution is shown at the timestamp $t=0.12 \, \mathrm{s}$ for three different meshes.}
  \label{fig:Burgers_adaptation_elements}
\end{figure}

We perform two additional simulations where we change the number of elements in the mesh (to 4 and 16 elements), and perform RL adaptation. Results are shown in Figure~\ref{fig:Burgers_adaptation_elements}. We observe that since the elements are transformed into the computational domain, where $\xi \in [-1,1]$, the model can perform the adaptation for an arbitrary size of the elements. If the number of elements is increased, the local gradients within each element have smaller fluctuations, and the RL agent selects a lower polynomial order. As expected, if a very coarse mesh is used, the agent selects a very high polynomial order to capture the solution, and if the mesh is fine, the agent decreases the polynomial order to reduce the cost of the simulation, while providing an accurate result.
%
%
%
%

\begin{figure}[h]
    \centering
    \includegraphics[width=0.5\textwidth]{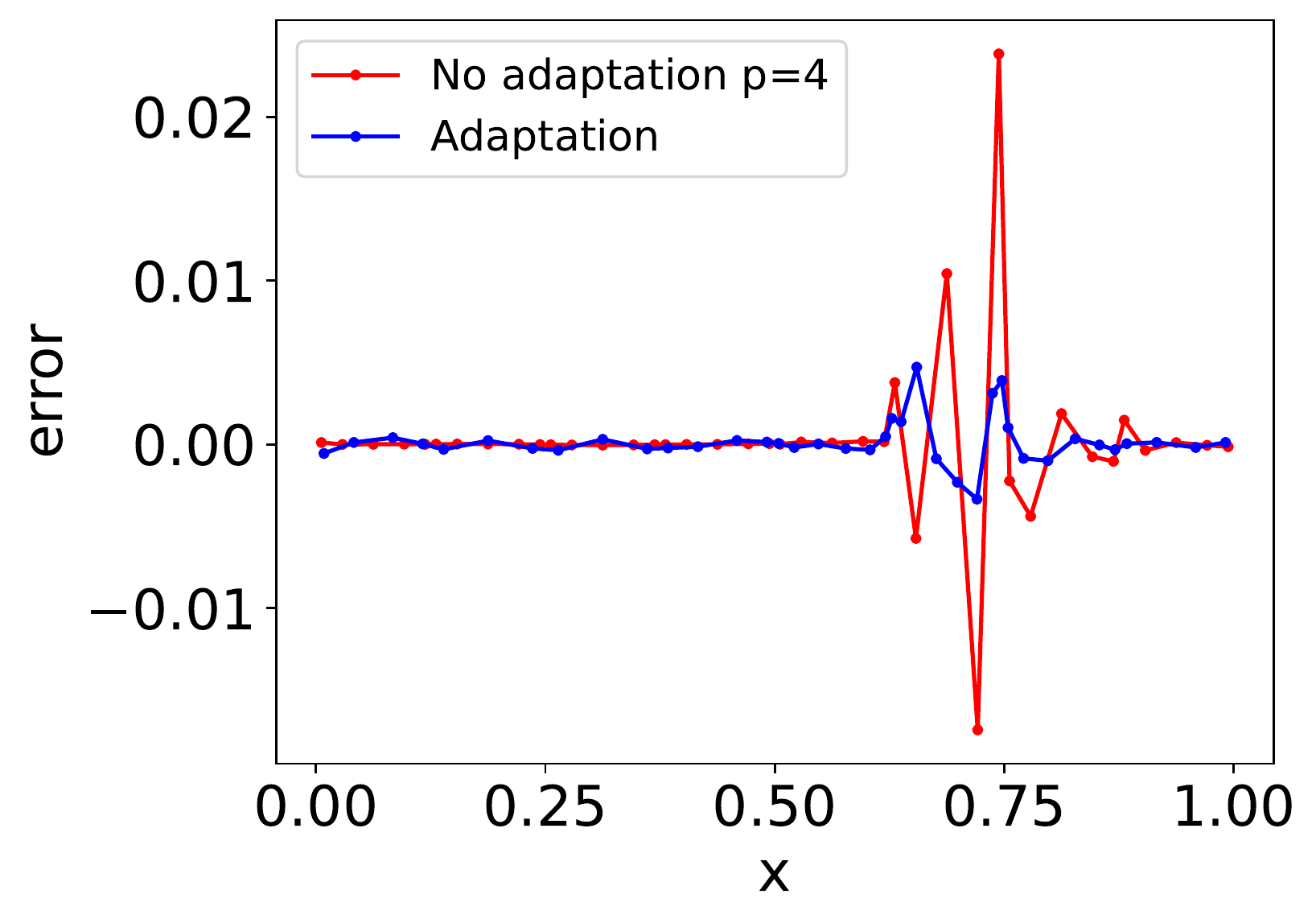}
    
    \caption{Point-wise error between the analytical solution and the numerical simulation with and without p-adaptation at the simulation time $t=0.12 \, \mathrm{s}$. The computational mesh has 8 elements.}
    \label{fig:Error}
\end{figure}

Finally, for a medium mesh of 8 elements, we use the analytical solution (Eq. \eqref{eq:Burgers_analytical_solution}) to  compute the point-wise errors across the domain, for the RL p-adaptation and the uniform $p=4$ simulations. These errors are calculated by subtracting the analytical solution from the numerical approximation at the Gauss nodes, where the solution is known. The results are shown in Figure~\ref{fig:Error}.
We see that the error is significantly smaller when using RL adaptation than when using a constant polynomial order $p=4$.
The solution not only shows improved accuracy, but also presents a more cost-effective approach, as evidenced by the data provided in Table~\ref{tab:computational_time}, which can be found in the next section.

In summary, we have shown that the RL adaptation is able to select the correct polynomial order in each element in the mesh. In the next section, we analyse the computational cost associated to the RL strategy.

\newpage
\subsection{Trade-off between accuracy and computational time}\label{tradeoff}
Previous results have shown that the RL p-adaptation algorithm is effective and capable of performing dynamic mesh adaptation, increasing the accuracy in problematic regions while reducing the polynomials (and the cost) in very smooth regions. 

However, there is no universal rule that establishes the best polynomial order for each element. This decision lies on the users, who should be able to adapt the algorithm to their own needs. In some cases, a better accuracy could be desired to capture very small fluctuations. In other cases, if the resources are more limited, it could be desirable to obtain a solution accurate enough, but prioritising the reduction of the computational time.

To have a trade-off between these two key points, a parameter, $\delta$, is introduced. This parameter is independent of the training of the RL agent and is included afterward to offer the user additional control. We can discriminate among three different cases:
\begin{enumerate}
    \item If $\delta=1$, the model shows the standard behaviour.
    \item If $0 < \delta < 1$, then the model tries to reduce the computational time by selecting a lower polynomial order during the adaptation process.
    \item If $\delta>1$, then the model tries to improve the accuracy by selecting a higher polynomial order during the adaptation process.
\end{enumerate}

This control parameter modifies the state by multiplying the error $e$ (see Section \ref{state}), to make the agent believe that the accuracy is higher (if $0 < \delta < 1$) or smaller (if $\delta>1$) than the real value. Therefore, the agent adapts the output to this new state to obtain a different set of solutions depending on the value of $\delta$.

\begin{figure}[htbp]
  \centering
  \begin{subfigure}[b]{0.4\textwidth}
    \includegraphics[width=\textwidth]{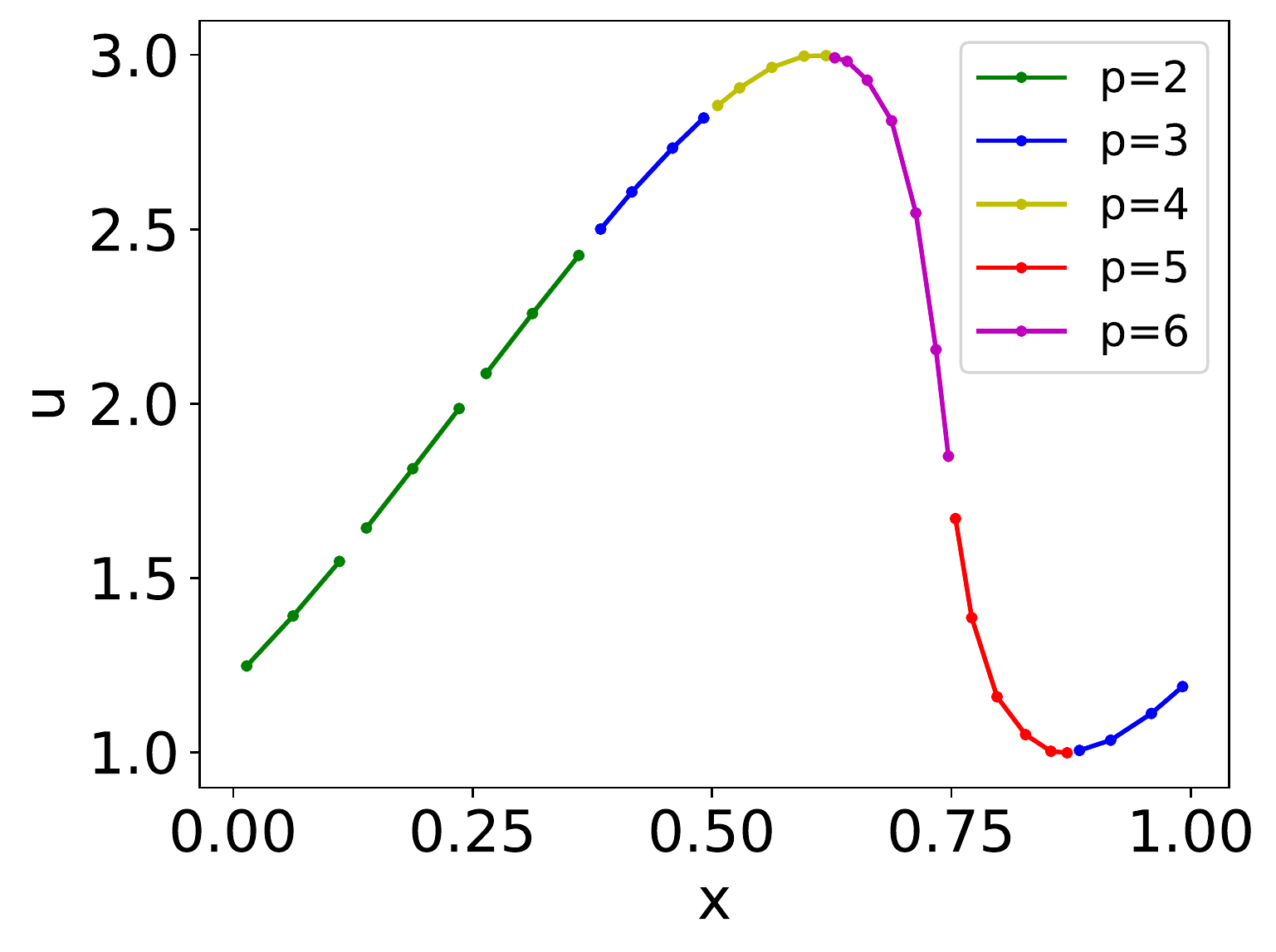}
    \caption{$\delta = 0.3$}
    \label{fig:adaptation_t0.12_delta0.3}
  \end{subfigure}
%
  %
  \begin{subfigure}[b]{0.4\textwidth}
    \includegraphics[width=\textwidth]{Adaptation_p4_t0.1200.pdf}
    \caption{$\delta = 1.0$}
    \label{fig:adaptation_t0.12_delta1.0}
  \end{subfigure}
%
  %
  \begin{subfigure}[b]{0.4\textwidth}
    \includegraphics[width=\textwidth]{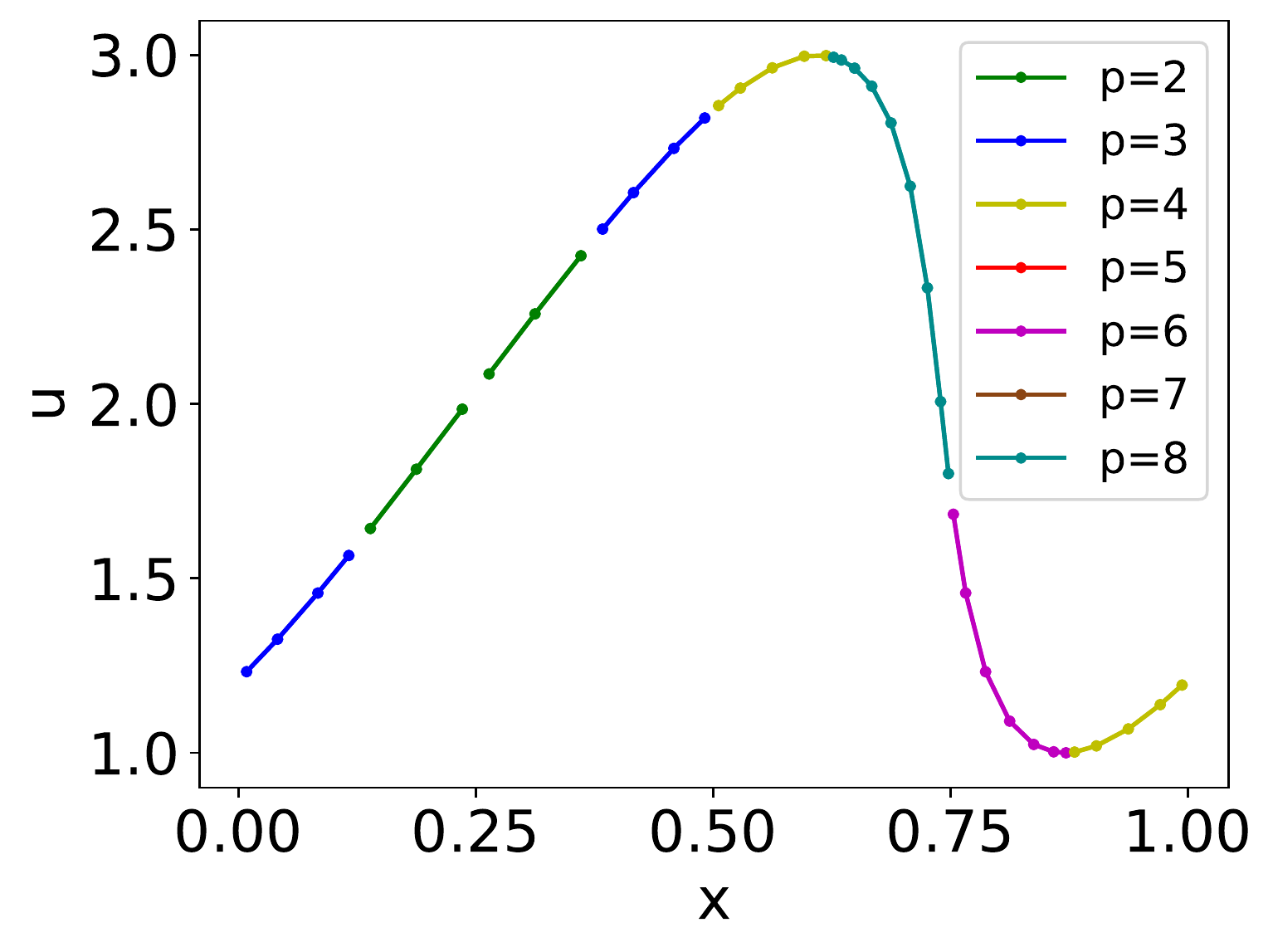}
    \caption{$\delta = 3.0$}
    \label{fig:adaptation_t0.12_delta3.0}
  \end{subfigure}
  \caption{1D inviscid Burgers' equation solved with a DGSEM spatial scheme and an explicit Euler temporal scheme, using a RL p-adaptation algorithm. The polynomial of each element of the computational mesh has been adapted once every $0.01 \, \mathrm{s}$ of simulation time, starting with a polynomial order $p=4$ in every element. The solution is shown at the timestamp $t=0.12 \, \mathrm{s}$ for three different values of the control parameter $\delta$.}
  \label{fig:Burgers_adaptation_delta}
\end{figure}

We simulate again using the RL adaptation and show the solution at the time $t=0.12 \, \mathrm{s}$ in Figure~\ref{fig:Burgers_adaptation_delta}, where three different values of $\delta$ are reported. In this case, it is clear that the general polynomial order has its value increased when $\delta$ has a higher value.

\begin{figure}[h]
    \centering
    \begin{subfigure}[b]{0.37\textwidth}
        \includegraphics[width=\textwidth, height = 4cm]{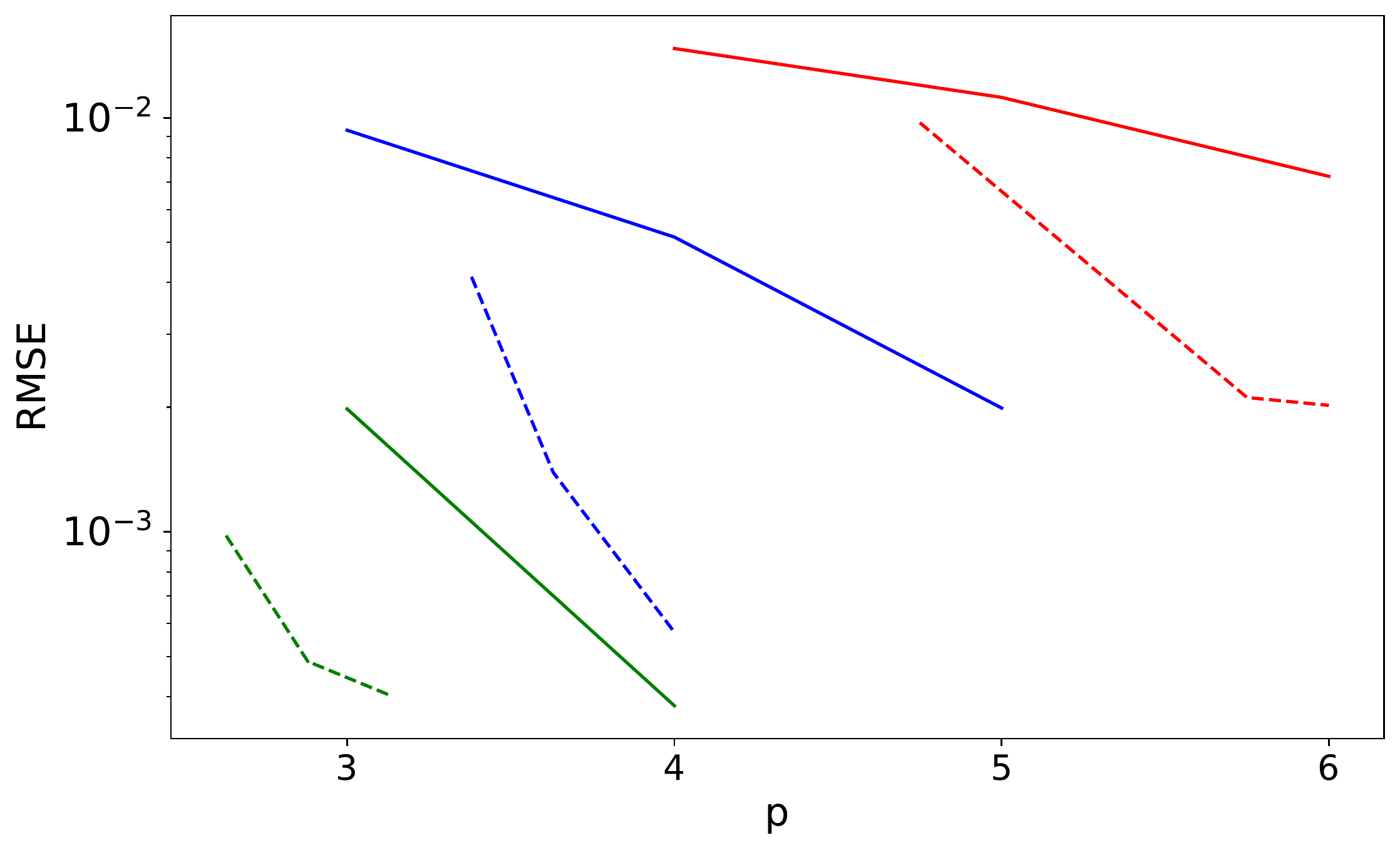}
        \caption{Error convergence.}
        \label{fig:Error_convergence}
      \end{subfigure}
    \quad
        \begin{subfigure}[b]{0.59\textwidth}
        \includegraphics[width=\textwidth, height = 4cm]{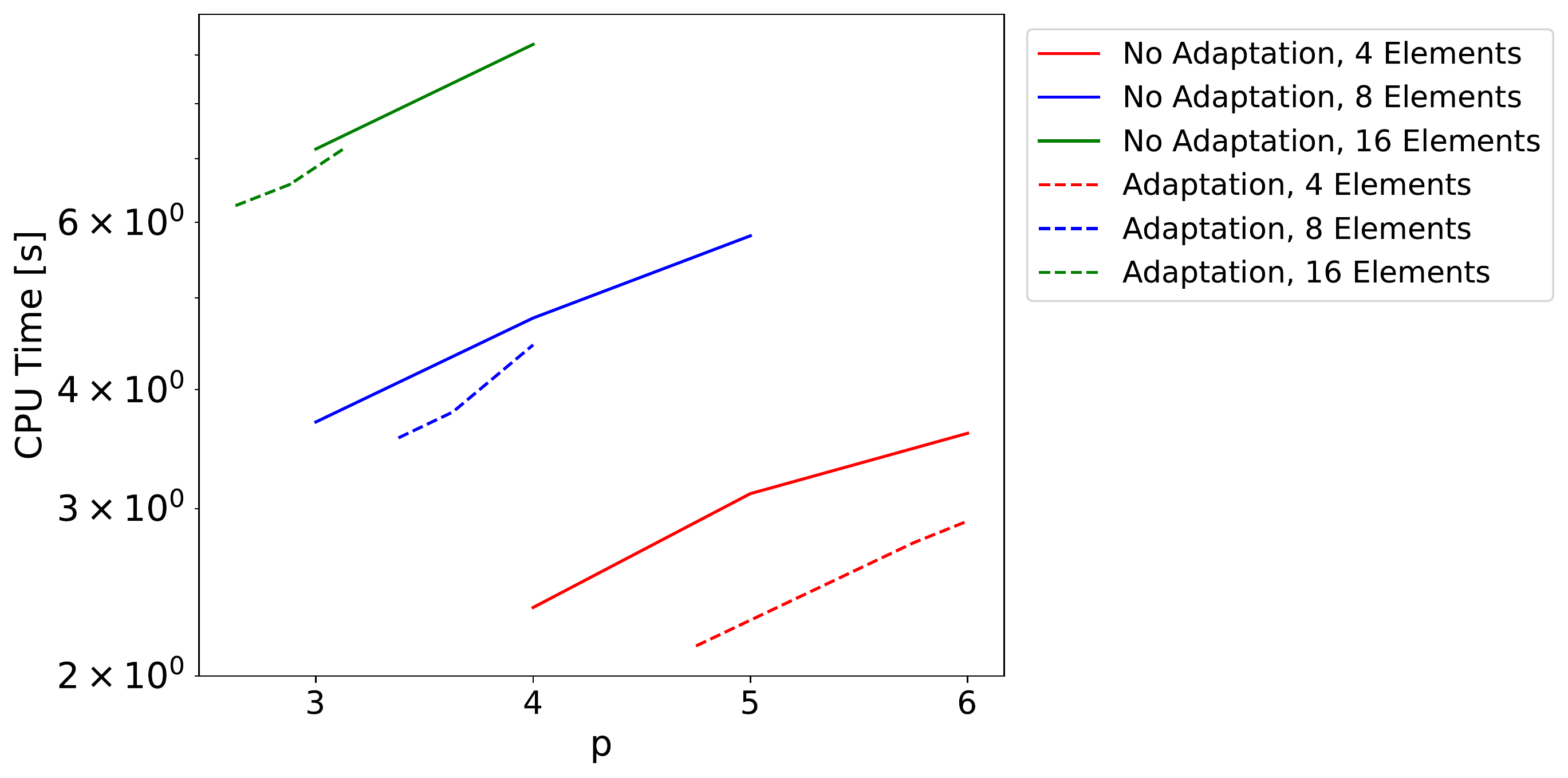}
        \caption{CPU time convergence.}
        \label{fig:cpu_time_convergence}
      \end{subfigure}
    
    \caption{Error $rmse$ and CPU time convergence with and without p-adaptation for three different meshes, as a function of the polynomial order $p$. When p-adaptation is used, the average polynomial order, $p_{av}$, is used, which represents the mean of the polynomials among the elements of the computational mesh.}
    \label{fig:error_CPUTime_convergence}
\end{figure}

To check the effect of the p-adaptation algorithm and the value of $\delta$ on the computational time and the accuracy of the solution, several simulations have been carried out and the results are summarised in Tables~\ref{tab:computational_time}, \ref{tab:error} and \ref{tab:ratio_efficiency_accuracy}. Table~\ref{tab:computational_time} shows that, when using 8 and 16 elements, the RL adaptation provides clear accelerations when compared to a uniform polynomial with $p=4$, for all values of $\delta$.
Even for high values of $\delta$ (e.g. $\delta=3.0$), the computational time using p-adaptation is similar or smaller than the time required to compute the simulation with a constant polynomial order. For completeness, Table~\ref{tab:error} shows a comparison of the $rmse$ of the approximate solution (in relation to the analytical solution provided in Eq. \eqref{eq:Burgers_analytical_solution}) with and without p-adaptation, where we see that improved accuracy is obtained in most cases when RL adaptation is used. 
We perform additional simulations without p-adaptation for different polynomial orders, and include the errors and computational times in Table~\ref{tab:computational_time_no_adaptation} and \ref{tab:error_no_adaptation}. These results have been compared with those obtained using p-adaptation, as shown in Figure~\ref{fig:error_CPUTime_convergence}, where it is clear that the RL p-adaptation approach improves the accuracy and reduces the computational cost in all cases.

Finally and for completness, Table~\ref{tab:ratio_efficiency_accuracy} shows a measurement of the trade-off between accuracy and efficiency. This quantity, which has a smaller value if the performance is better, has been obtained by multiplying the computational time (see Table~\ref{tab:computational_time}) by the error (see Table~\ref{tab:error}). In th table, we have highlighted in bold the best results for each mesh. Based on these results, the value of the control parameter, $\delta$, should be increased if the mesh is refined to obtain accurate and fast solutions.
In general, the proposed method provides a good trade-off between efficiency and accuracy, and both are improved in most cases.

\begin{table}[h]
    \centering
    \begin{tabular}{|C{1.5cm}|C{2.5cm}|C{2.5cm}|C{2.5cm}|C{2.5cm}|}
    \hline
    \textbf{Elements} & \textbf{No adaptation $p=3$} & \textbf{No adaptation $p=4$} & \textbf{No adaptation $p=5$} & \textbf{No adaptation $p=6$} \\ \hline
    4 & $1.89 \, \mathrm{s}$ & $2.36 \, \mathrm{s}$ & $3.11 \, \mathrm{s}$ & $3.60 \, \mathrm{s}$ \\ \hline
    8 & $3.70 \, \mathrm{s}$ & $4.76 \, \mathrm{s}$ & $5.81 \, \mathrm{s}$ & $6.93 \, \mathrm{s}$ \\ \hline
    16 & $7.17 \, \mathrm{s}$ & $9.24 \, \mathrm{s}$ & $11.33 \, \mathrm{s}$ & $14.27 \, \mathrm{s}$ \\ \hline
    \end{tabular}
    \caption{Comparison between the computational time without p-adaptation for three meshes and four values of the polynomial order $p$.}
    \label{tab:computational_time_no_adaptation}
\end{table}

\begin{table}[h]
    \centering
    \begin{tabular}{|C{1.5cm}|C{2.5cm}|C{3.0cm}|C{3.0cm}|C{3.0cm}|}
    \hline
    \textbf{Elements} & \textbf{No adaptation $p=4$} & \textbf{Adaptation $\delta=0.3$} & \textbf{Adaptation $\delta=1.0$} & \textbf{Adaptation $\delta=3.0$}  \\ \hline
    4 & $2.36 \, \mathrm{s}$ & $2.15 \, \mathrm{s}$ ($p_{av}=4.75$) & $2.76 \, \mathrm{s}$ ($p_{av}=5.75$) & $2.91 \, \mathrm{s}$ ($p_{av}=6.0$)\\ \hline
    8 & $4.76 \, \mathrm{s}$ & $3.56 \, \mathrm{s}$ ($p_{av}=3.38$) & $3.79 \, \mathrm{s}$ ($p_{av}=3.63$) & $4.46 \, \mathrm{s}$ ($p_{av}=4.0$)  \\ \hline
    16 & $9.24 \, \mathrm{s}$ & $6.25 \, \mathrm{s}$ ($p_{av}=2.63$) & $6.58 \, \mathrm{s}$ ($p_{av}=2.88$)  & $7.18 \, \mathrm{s}$ ($p_{av}=3.13$) \\ \hline

    \end{tabular}
    \caption{Comparison between the computational time with and without p-adaptation for three meshes and three values of the control parameter $\delta$. The time required to adapt the mesh for each $0.01 \, \mathrm{s}$ is included in these values. The average polynomial order, $p_{av}$, represents the mean of the polynomials among the elements of the computational mesh.}
    \label{tab:computational_time}
\end{table}

\begin{table}[h]
    \centering
    \begin{tabular}{|C{1.3cm}|C{2.5cm}|C{2.5cm}|C{2.5cm}|C{2.5cm}|}
    \hline
    \textbf{Elements} & \textbf{No adaptation $p=3$} & \textbf{No adaptation $p=4$} & \textbf{No adaptation $p=5$} & \textbf{No adaptation $p=6$}  \\ \hline
    4 & $1.74 \cdot 10^{-2}$ & $1.47 \cdot 10^{-2}$ & $1.12 \cdot 10^{-2}$ & $7.22 \cdot 10^{-3}$ \\ \hline
    8 & $9.33 \cdot 10^{-3}$ & $5.15 \cdot 10^{-3}$ & $1.99 \cdot 10^{-3}$ & $6.59 \cdot 10^{-4}$ \\ \hline
    16 & $1.98 \cdot 10^{-3}$ & $3.80 \cdot 10^{-4}$ & $2.98 \cdot 10^{-4}$ & $2.25 \cdot 10^{-4}$ \\ \hline
    \end{tabular}
    \caption{Comparison between the $rmse$ of the approximated solution in relation to the analytical solution without p-adaptation for three meshes and four values of the polynomial order $p$.}
    \label{tab:error_no_adaptation}
\end{table}

\begin{table}[h]
    \centering
    \begin{tabular}{|C{1.3cm}|C{2.3cm}|C{3.3cm}|C{3.3cm}|C{3.3cm}|}
    \hline
    \textbf{Elements} & \textbf{No adaptation $p=4$} & \textbf{Adaptation $\delta=0.3$} & \textbf{Adaptation $\delta=1.0$} & \textbf{Adaptation $\delta=3.0$}  \\ \hline
    4 & $1.47 \cdot 10^{-2}$ & $9.73 \cdot 10^{-3}$ ($p_{av}=4.75$) & $2.11 \cdot 10^{-3}$ ($p_{av}=5.75$) & $2.02 \cdot 10^{-3}$ ($p_{av}=6.0$)  \\ \hline
    8 & $5.15 \cdot 10^{-3}$ & $4.13 \cdot 10^{-3}$ ($p_{av}=3.38$) & $1.39 \cdot 10^{-3}$ ($p_{av}=3.63$) & $5.72 \cdot 10^{-4}$ ($p_{av}=4.0$)  \\ \hline
    16 & $3.80 \cdot 10^{-4}$ & $9.79 \cdot 10^{-4}$ ($p_{av}=2.63$) & $4.86 \cdot 10^{-4}$ ($p_{av}=2.88$) & $4.03 \cdot 10^{-4}$ ($p_{av}=3.13$)  \\ \hline
    

    \end{tabular}
    \caption{Comparison between the $rmse$ of the approximated solution in relation to the analytical solution with and without p-adaptation for three meshes and three values of the control parameter $\delta$. The average polynomial order, $p_{av}$, represents the mean of the polynomials among the elements of the computational mesh.}
    \label{tab:error}
\end{table}

\begin{table}[h]
    \centering
    \begin{tabular}{|C{1.3cm}|C{2.3cm}|C{3.3cm}|C{3.4cm}|C{3.4cm}|}
    \hline
    \textbf{Elements} & \textbf{No adaptation $p=4$} & \textbf{Adaptation $\delta=0.3$} & \textbf{Adaptation $\delta=1.0$} & \textbf{Adaptation $\delta=3.0$}  \\ \hline
    4 & $3.47 \cdot 10^{-2}$ & $2.09 \cdot 10^{-2}$ ($p_{av}=4.75$) & $\mathbf{5.82 \cdot 10^{-3}}$ ($p_{av}=5.75$) & $5.88 \cdot 10^{-3}$ ($p_{av}=6.0$)  \\ \hline
    8 & $2.45 \cdot 10^{-2}$ & $1.47 \cdot 10^{-2}$ ($p_{av}=3.38$) & $5.27 \cdot 10^{-3}$ ($p_{av}=3.63$) & $\mathbf{2.55 \cdot 10^{-3}}$ ($p_{av}=4.0$)  \\ \hline
    16 & $3.51 \cdot 10^{-3}$ & $6.12 \cdot 10^{-3}$ ($p_{av}=2.63$) & $3.20 \cdot 10^{-3}$ ($p_{av}=2.88$) & $\mathbf{2.89 \cdot 10^{-3}}$ ($p_{av}=3.13$)  \\ \hline
    \end{tabular}
    \caption{Measurement of the trade-off between efficiency (the computational time) and accuracy (the error $rmse$), by multiplying both quantities, with and without p-adaptation for three meshes and three values of the control parameter $\delta$. The best result (the smallest) for each mesh is highlighted in bold. The average polynomial order, $p_{av}$, represents the mean of the polynomials among the elements of the computational mesh.}
    \label{tab:ratio_efficiency_accuracy}
\end{table}

\newpage 
The p-adaptation stratgy shows good performance and appears competitive for speeding up numerical simulations. It allows to obtain a significant reduction of the computational cost while maintaining a high level of accuracy. 

\newpage
\section{Conclusions}\label{conc}
In conclusion, reinforcement learning (RL) is a powerful approach to automate and optimise mesh adaptation in the field of high-order numerical simulations. By integrating RL algorithms into high-order simulations, we have demonstrated the ability of RL agents to learn and adapt the computational mesh based on evolving solutions.

The application of RL in mesh p-adaptation brings numerous benefits. First, it reduces the reliance on manual intervention, enabling automated and adaptive mesh refinement. RL agents can discover non-obvious mesh modifications that enhance the accuracy while reducing cost. The agent's ability to learn optimal mesh modifications based on the solution characteristics allows for the discovery of innovative strategies that enhance simulation accuracy and efficiency.
Despite these advances, challenges remain in the application of RL for mesh adaptation. Sample inefficiency, exploration in high-dimensional spaces, and generalisation to unseen scenarios are among the key hurdles that must be addressed. 
By leveraging RL's ability to learn optimal mesh modifications, we can enhance the efficiency of numerical simulations while reducing human intervention.

\section*{Acknowledgments}
EF would like to thank the support of
Agencia Estatal de Investigación (for the grant "Europa Excelencia 2022" Proyecto EUR2022-134041/AEI/10.13039/501100011033) y del Mecanismo de Recuperación y Resiliencia de la Unión Europea, and the Comunidad de Madrid and Universidad Politécnica de Madrid for the Young Investigators award: APOYO-JOVENES-21-53NYUB-19-RRX1A0. 
Finally, all authors gratefully acknowledge the Universidad Politécnica de Madrid (www.upm.es) for providing computing resources on Magerit Supercomputer.

\appendix

\section{DGSEM}
\label{app:dgsem}
To derive the DGSEM, the physical domain is tessellated with non-overlapping elements, $e$, which are geometrically transformed into a reference element $el=[-1,1]$. This transformation is carried out using an affine mapping that relates the physical coordinates $x$ and the local reference coordinates $\xi$. The transformation is applied to the Burgers' equation \eqref{eq:Burgers_equation}, resulting in the following:

\begin{equation}
J u_t  + \frac{\partial f}{\partial \xi} = 0,
\label{eq:burgers_transformed}
\end{equation}
where $J$ is the Jacobian of the mapping, and $f=\frac{u^2}{2}$ is the flux. 
Next, we multiply Eq. \eqref{eq:burgers_transformed} by a element-wise smooth test function $\phi_j$ (i.e., in our case a polynomial), for $0\leq j\leq p$, where $p$ is the polynomial degree, and integrate over an element $el$ to obtain the weak form

%
\begin{equation}\label{eq::NS2}
\int_{-1}^1J u_t\phi_j+\int_{-1}^1 \frac{\partial f}{\partial \xi}\phi_j  =0.
\end{equation}
We can now integrate by parts the second integral, to obtain a local weak form of the equations (one per element) with the boundary fluxes separated from the interior
\begin{equation}
\label{eq:DG1DWeak_reference_approx_4}
\int_{-1}^1 J u_t \phi_j+ \left.f\phi_j\right|_{-1}^1-\int_{-1}^1 f \frac{\partial \phi_j}{\partial \xi}  =0.
\end{equation}
We replace discontinuous fluxes at inter--element faces by a numerical flux, $f^{\star}$, to couple neighbouring elements, 
\begin{equation}
\label{eq:DG1DWeak_reference_approx_4}
\int_{-1}^1 J u_t \phi_j+ \left.f^{\star}\phi_j\right|_{\partial el}-\int_{-1}^1 f \frac{\partial \phi_j}{\partial \xi}  =0.
\end{equation}
In this work, we have chosen the Roe numerical flux for $f^{\star}$.
The final step to obtain a usable numerical scheme is to approximate the numerical solution, fluxes, and test functions by polynomials of order $p$, and to use Gaussian quadrature rules to numerically approximate the integrals. Here, we use for Gauss-Legendre quadrature points. The resulting system of ordinary differential equations can be solved by any explicit or implicit time-stepping method; in this work we use an explicit Euler scheme. For more details, see \cite{koprivaImplementingSpectralMethods2009,FERRER2023108700}. 

\bibliographystyle{plain}
\bibliography{sample}

\end{document}